\documentclass[11pt]{article}

\usepackage{graphicx}
\usepackage{subfigure}
\usepackage[cmex10]{amsmath}
\usepackage{amssymb}
\usepackage{amsfonts}
\usepackage{url}
\usepackage{bm}
\usepackage{multirow}
\usepackage{array}
\usepackage{algorithm}
\usepackage{algorithmic}
\usepackage{natbib}
\usepackage[margin=1.25in]{geometry}
\usepackage{authblk}

\usepackage{pgfplots}
\pgfplotsset{compat=newest}

\newtheorem{theorem}{Theorem}
\newtheorem{corollary}{Corollary}
\newtheorem{lemma}{Lemma}

\newtheorem{proposition}{Proposition}

\newtheorem{problem}{Problem}

\renewcommand{\matrix}[1]{\ensuremath{\begin{bmatrix} #1 \end{bmatrix}}}
\newcommand{\tr}[1]{\ensuremath{\mathrm{tr} \! \left[ #1 \right] \! }}
\newcommand{\argmin}{\mathop{\mathrm{argmin}}\limits}
\newcommand{\argminl}{\mathop{\mathrm{argmin}}\nolimits}

\newcommand{\R}{\ensuremath{\mathbb{R}}}

\newcommand{\vi}{{\vert\kern-0.25ex\vert\kern-0.25ex\vert}}
\newcommand{\comp}{{\ensuremath{\rm c}}}

\makeatletter
\newcommand{\figcaption}[1]{\def\@captype{figure}\caption{#1}}
\newcommand{\tblcaption}[1]{\def\@captype{table}\caption{#1}}
\makeatother

\begin{document}

\title{Consistent Nonparametric Different-Feature Selection\\via the Sparsest $k$-Subgraph Problem}

\author[1,2]{Satoshi Hara\thanks{satohara@nii.ac.jp}}
\author[3]{Takayuki Katsuki}
\author[3]{Hiroki Yanagisawa}
\author[4]{Masaaki Imaizumi}
\author[5]{Takafumi Ono}
\author[6]{Ryo Okamoto}
\author[6]{Shigeki Takeuchi}
\affil[1]{National Institute of Informatics, Japan}
\affil[2]{JST, ERATO, Kawarabayashi Large Graph Project, Japan}
\affil[3]{IBM Research -- Tokyo, Japan}
\affil[4]{The Institute of Statistical Mathematics, Japan}
\affil[5]{University of Bristol, UK}
\affil[6]{Kyoto University, Japan}

\date{}


\maketitle

\begin{abstract}%
Two-sample feature selection is the problem of finding features that describe a difference between two probability distributions, which is a ubiquitous problem in both scientific and engineering studies.
However, existing methods have limited applicability because of their restrictive assumptions on data distributoins or computational difficulty.
In this paper, we resolve these difficulties by formulating the problem as a sparsest $k$-subgraph problem.
The proposed method is nonparametric and does not assume any specific parametric models on the data distributions.
We show that the proposed method is computationally efficient and does not require any extra computation for model selection.
Moreover, we prove that the proposed method provides a consistent estimator of features under mild conditions.
Our experimental results show that the proposed method outperforms the current method with regard to both accuracy and computation time.
\end{abstract}

\section{Introduction}
\label{sec:intro}

Two-sample feature selection is the task of finding features with distribution differences between two datasets.
Feature selection helps us understand what causes differences between datasets, which is a fundamental problem in both scientific and engineering studies.
Important example tasks include the two-sample test~\citep{benjamini1995controlling,gretton2012kernel,mueller2015principal} and anomaly detection~\citep{taguchi2000new,ide2009proximity,hara2015consistent}.
For example, in gene expression data analysis, a two-sample test-based approach allows us to find genes that are specific to some subtypes~\citep{mueller2015principal}.
In the anomaly detection context, one can find causes of an error by localizing features that behave differently between datasets sampled before and after the occurrence of the error~\citep{ide2009proximity,hara2015consistent}.

In this paper, we focus on finding features that describe a difference between two probability distributions.
Suppose we have independent and identically distributed (i.i.d.) samples from probability distributions $p(\bm{x})$ and $q(\bm{x})$ of sizes $N$ and $M$, respectively, where $\bm{x} \in \R^D$ is a $D$-dimensional feature.
Here, without loss of generality, we assume $N \ge M$ throughout this paper. 
Using these samples, we aim to find a subset of features $S \subseteq \{1, 2, \ldots, D\}$ for which the two distributions do not match.
Intuitively, we expect that $p(\bm{x}_S) \neq q(\bm{x}_S)$ and $p(\bm{x}_{S^\comp}) = q(\bm{x}_{S^\comp})$ hold, where $\bm{x}_S$ and $\bm{x}_{S^\comp}$ denote subsets of a random variable $\bm{x}$ specified by the set $S$ and its complement $S^\comp$, respectively.
We refer to this problem as \emph{different-feature selection}.

There have been several studies on different-feature selection in the two sample test and anomaly detection contexts.
In the two sample test context, \cite{benjamini1995controlling} proposed comparing each single feature using statistical tests and then adjusting the false discovery rate using the Bonferroni method~\citep{bonferroni1936teoria}.
In the anomaly detection context, in which the objective is to find features with anomalies, the Mahalanobis-Taguchi System (MT)~\citep{taguchi2000new} is one of the most classic methods.
The MT models both $p$ and $q$ as Gaussians and then finds features with different means or covariances.
Following MT, several lines of research have focused on different-feature selection under the Gaussian setting.
\cite{hirose2009network} proposed using the change in inter-sensor correlations to find features with distribution changes.
\cite{jiang2011anomaly} proposed a PCA-based method.
\cite{ide2007computing,ide2009proximity} used the changes in correlation and partial correlation.
In our previous study~\citep{hara2015consistent}, we proposed an algorithm with a consistency guarantee.

Unlike Gaussian-based methods, only a little has been studied about nonparametric different-feature selection methods.
The first nonparametric different-feature selection method, called SPARDA, was proposed by \cite{mueller2015principal}. 
SPARDA finds a feature set $S$ by searching for a subspace with the maximum distribution difference by solving a nonconvex problem.
In particular, \cite{mueller2015principal} used a nonparametric metric called the Wasserstein distance~\citep{gibbs2002choosing} to measure the difference between the distributions.
Because the Wasserstein distance is nonparametric, SPARDA does not assume any specific parametric models on $p$ and $q$.
This property contrasts with MT and its variants, which use the Gaussian distributions.
This nonparametric nature of SPARDA is favorable in practice because we usually do not know the data distribution models, and they can be non-Gaussian in many cases.
\cite{mueller2015principal} also proved that SPARDA provides a consistent estimator of the feature subset $S$.
The major difficulty with SPARDA, however, is solving the nonconvex optimization problem.
The authors proposed a \textit{relax and tighten} procedure that can find nearly global optima; however, this procedure leads to high computational complexity.
It solves a semidefinite program at every iteration, which runs in $O(D^3 N^2)$ time.
Therefore, applying the relax and tighten procedure to large datasets is difficult.
Projected gradient ascent is a faster alternative method that runs in $O(DN + N \log N)$ time per iteration.
However, it is easily trapped by local optima, as we demonstrate in our experiments.
Note that in practice, the computation time of these methods is further increased by the need for cross validation for model selection; SPARDA needs to choose an optimal regularization parameter.

This literature survey reveals the limitations of existing different-feature selection methods.
The Gaussian-based methods have limited applicability due to the restrictive Gaussian assumption, whereas the nonparametric SPARDA approach has computational difficulty.
These limitations hinder us from studying the causes of differences in large complex datasets.
Therefore, a computationally efficient different-feature selection method with an assumption that is less restrictive than that of current methods is required to fulfill our practical needs.

In this paper, we propose a simple nonparametric method for different-feature selection that resolves these two problems, namely, restrictive assumptions and computational inefficiency, by extending our preliminary study~\citep{hara2017consistent}.
The current paper differs from our preliminary study in two ways.
First, the analysis of the computational complexity is improved; in our preliminary study, only the average time complexity was evaluated.
In this study, we derive the improved worst case complexity.
Second, the feature selection consistency theorem is improved; in our preliminary study, only the asymptotic setting was studied.
Here,  we study the finite sample case.
These differences come from the modification of the proposed method.
While we used KL-divergence as the difference metric between the distributions in our preliminary study, we now replace it with a modified Kolmogorov-Smirnov (KS) statistic, which we describe in detail in Section~\ref{sec:proposed}.

In summary, our major contributions are twofold.
First, we propose a simple nonparametric method for different-feature selection.
The proposed method does not assume any specific parametric models on $p$ and $q$, and its time complexity is only $O(D^2 L N \log N)$ where $L$ is an algorithm parameter.
Moreover, the proposed method does not require the optimization of any regularization parameters; thus, it does not require any extra computation for model selection.
We formulate the problem as a sparsest $k$-subgraph problem~\citep{watrigant2016approx} using the KS statistic.
Although the problem is NP-hard in general, we derive a nearly global optimum solution using a greedy method.

Second, we provide a feature selection consistency theorem for the proposed method.
Although there are several studies regarding different-feature selection, only a couple of studies give consistency guarantees~\citep{hara2015consistent,mueller2015principal}.
Our theoretical result shows that the probability of the misspecification of the feature set decays exponentially as the number of samples $N$ and $M$ increase.
Unlike the Gaussian-based method~\citep{hara2015consistent}, we prove that this guarantee holds even under non-Gaussian settings without assuming any specific distribution models on $p$ and $q$.
Our consistency guarantee requires conditions only on the KS statistic between the data distributions but not on their distribution models.
Moreover, the result shows that for the probability of the misspecification to be smaller than $\epsilon$, $N \approx O\left(\max\left\{k^4/\eta^2, k^2/\eta\right\} \log D / \epsilon \right)$ samples suffice, where $\eta$ and $k$ are problem dependent parameters.

Our experimental results confirm the high accuracy and computational efficiency of the proposed method for both synthetic and real-world data.
We found that the proposed nonparametric method can detect a complex distribution difference effectively and outperforms Gaussian-based methods.
We also compared the proposed method and SPARDA with projected gradient ascent for both accuracy and runtime.
The results show that the proposed method attains higher accuracy in many cases.
We conjecture that SPARDA tends to be trapped by local optima, whereas the proposed method is able to find nearly global optima using the greedy method.
We also observed that the speed of the proposed method is comparable to or even several times faster than that of SPARDA.

\paragraph{Notation:}
Let $[D] := \{ 1, 2, \ldots, D \}$ for $D \in \mathbb{N}$.
For a vector $\bm{x} \in \R^D$, $x_d$ is its $d$-th component, and for a matrix $H \in \R^{D \times D}$, $H_{ij}$ is its $(i, j)$-th component.
For a set $S \subseteq [D]$, $S^\comp := [D] \setminus S$ is its complement.
For a vector $\bm{x}$ and a set $S \subseteq [D]$, $\bm{x}_S := \{x_d \mid d \in S \}$ is a feature subset.
Moreover, $\mathcal{N}(\bm{\mu}, \Sigma)$ denotes the Gaussian distribution with mean $\bm{\mu}$ and covariance $\Sigma$.
$\mathcal{U}(\alpha, \beta)$ denotes the uniform distribution in $[\alpha, \beta] \subset \mathbb{R}$.
$\bm{0}_D$ and $\bm{1}_D$ denote $D$-dimensional vectors with all entries equal to zero and one, respectively.
For a statement $a$, $\mathbb{I}(a)$ denotes the indicator of $a$, i.e., $\mathbb{I}(a) = 1$ if $a$ is true, and $\mathbb{I}(a) = 0$ if $a$ is false.
For a function $f$, we write the supremum norm as $\| f \|_\infty := \sup_x |f(x)|$.

\section{Preliminaries}
\label{sec:preliminary}

Across the paper, we use the Kolmogorov-Smirnov (KS) statistic~\citep{hollander2013nonparametric} as the basic measurement of the difference between the two distributions.
We therefore start by reviewing the KS statistic, one of the most popular nonparametric two-sample test statistics.
The KS statistic is used to verify whether two distributions are different.
Suppose the two random variables $y, z \in \R$ follow distributions $p(y)$ and $q(z)$, respectively.
Here, we also denote their distribution functions by $P(y) := \int_{-\infty}^y p(y') dy'$ and $Q(z) := \int_{-\infty}^z q(z') dz'$, respectively.
The KS statistic is defined using these distribution functions as
\begin{align}
	{\rm KS}(p, q) := \|P - Q\|_\infty ,
\end{align}
which is equivalent to the $L_\infty$-distance between the two distribution functions.
We note that the KS statistic is always bounded as ${\rm KS}(p, q) \in [0, 1]$ from its definition.
In practice, we do not know true distributions $p(y)$ and $q(z)$ or their distribution functions $P(y)$ and $Q(z)$.
Here, let the i.i.d.\ observations be $\mathcal{P} = \{y^{(n)}\}_{n=1}^N \overset{\rm i.i.d.}{\sim} p(y)$ and $\mathcal{Q} = \{z^{(m)}\}_{m=1}^M \overset{\rm i.i.d.}{\sim} q(z)$.
The empirical version of the KS statistic is given by
\begin{align}
	{\rm KS}(\hat{p}, \hat{q}) := \| \hat{P} - \hat{Q} \|_\infty ,
\end{align}
where $\hat{p}(y)$ and $\hat{q}(z)$ are empirical distributions, and $\hat{P}(y)$ and $\hat{Q}(z)$ are empirical distribution functions given by
\begin{align}
	\hat{P}(y) := \frac{1}{N} \sum_{n=1}^N \mathbb{I}(y \le y^{(n)}), \qquad \hat{Q}(z) := \frac{1}{M} \sum_{m=1}^M \mathbb{I}(z \le z^{(m)}) .
\end{align}
We note that, as shown in Algorithm~\ref{alg:ks}, the empirical KS statistic ${\rm KS}(\hat{p}, \hat{q})$ can be computed in $O(N \log N)$ time using sorting.

\begin{figure}[t]
\centering
\begin{algorithm}[H]
	\caption{Computing Empirical Kolmogorov-Smirnov Statistic}
	\label{alg:ks}
	\begin{algorithmic}
		\REQUIRE Datasets $\mathcal{P} = \{y^{(n)}\}_{n=1}^N$, $\mathcal{Q} = \{z^{(m)}\}_{m=1}^M$
		\ENSURE Empirical Kolmogorov-Smirnov statistic $h = {\rm KS}(\hat{p}, \hat{q})$
		\STATE $\mathcal{P} \leftarrow$ sort $\mathcal{P}$ in an ascending order
		\STATE $\mathcal{Q} \leftarrow$ sort $\mathcal{Q}$ in an ascending order
		\STATE $h \leftarrow 0$
		\STATE $i \leftarrow 1$
		\STATE $j \leftarrow 1$
		\WHILE{$i \le N$ and $j \le M$}
			\IF{$y^{(i)} = z^{(j)}$}
				\STATE $i \leftarrow i + 1$
				\STATE $j \leftarrow j + 1$
			\ELSIF{$y^{(i)} < z^{(j)}$}
				\STATE $i \leftarrow i + 1$
			\ELSE
				\STATE $j \leftarrow j + 1$
			\ENDIF
			\STATE $h \leftarrow \max \left\{ h, |(i-1)/N - (j-1)/M| \right\}$
		\ENDWHILE
	\end{algorithmic}
\end{algorithm}
\end{figure}

\section{Problem Definition}
\label{sec:problem}

Here, we define the different-feature selection problem considered in this paper.
Let $\bm{x} := (x_1, x_2, \ldots, x_D)^\top \in \R^D$ be a $D$-dimensional feature vector.
We aim to find features in which the distributions do not match between two distributions.
That is, for a subset $S^* \subseteq [D]$, we expect that there is a distribution difference in the $d$-th feature $x_{d}$ when $d \in S^*$, whereas there is no distribution difference in the $d'$-th feature $x_{d'}$ when $d' \notin S^*$.
We formalize the problem as follows.
\begin{problem}[Different-Feature Selection]
	\label{prob:anomaly}
	Given i.i.d.\ samples $\mathcal{P} = \{\bm{y}^{(n)}\}_{n=1}^{N} \overset{\rm i.i.d.}{\sim} p(\bm{x})$ and $\mathcal{Q} = \{\bm{z}^{(m)}\}_{m=1}^{M} \overset{\rm i.i.d.}{\sim} q(\bm{x})$, identify the set $S^* \subseteq [D]$ that satisfies
	\begin{align}
		p(\bm{x}_{{S^*}^\comp}) &= q(\bm{x}_{{S^*}^\comp}) , \label{eq:problem1} \\
		p(\bm{x}_{{S^*}^\comp \cup \{d\}}) &\neq q(\bm{x}_{{S^*}^\comp \cup \{d\}}) , \; \forall d \in S^* . \label{eq:problem2}
	\end{align}
\end{problem}
Here, we impose one technical assumption, which is that the feature set $S^*$ is uniquely identifiable; otherwise the problem is ill-posed.

Conditions (\ref{eq:problem1}) and (\ref{eq:problem2}) respectively require that the distributions match on feature subset ${S^*}^\comp$ but that this equation does not hold when feature $d \in S^*$ is removed from $S^*$ and added to ${S^*}^\comp$.

We note that Problem~\ref{prob:anomaly} is a generalization of a common feature selection problem for binary classification.
Altough existing methods, such as Lasso logistic regression~\citep{lee2006efficient}, search for discriminative features between the two classes, in Problem~\ref{prob:anomaly}, we also search for non-discriminative features with distribution differences (e.g., features with variance changes).

\section{Proposed Problem Formulation}
\label{sec:proposed}

We propose a simple nonparametric method for different-feature selection that satisfies two requirements, i.e., a less restrictive assumption and computational efficiency.
In this section, we formulate the different-feature selection problem as a sparsest $k$-subgraph problem~\citep{watrigant2016approx}, which leads to computationally efficient algorithms and desirable theoretical properties, which we describe in the upcoming sections.
Specifically, we formulate the problem by focusing only on the difference of the marginal distributions on the pair of features.
The proposed problem formulation can capture the differences of the higher-order moments of distributions, which is overlooked by the Gaussian-based methods.

\subsection{Different-Feature Selection as a Sparsest $k$-Subgraph Problem}

We formulate the different-feature selection problem as a sparsest $k$-subgraph problem using a matrix $\hat{H} \in \mathbb{R}_+^{D \times D}$, where each element of $\hat{H}$ represents the difference of the marginal distribution of the corresponding feature pair.
The proposed formulation is based on the assumption that matrix $\hat{H}$ leads to $\hat{S} = S^*$, where
\begin{align}
	\hat{S}^\comp := \argmin_{S^\comp \subseteq [D]} \sum_{i, j \in S^\comp} \hat{H}_{ij}, \; {\rm s.t.} \; |S^\comp| = k .
	\label{eq:kss}
\end{align}
Here, we assume that the size of ${S^*}^{\rm c}$ is known to be $k$.
We later describe how we design matrix $\hat{H}$.
Once we can design matrix $\hat{H}$, we can identify feature set $S^*$ by solving problem (\ref{eq:kss}), which is known as the sparsest $k$-subgraph problem~\citep{watrigant2016approx}.
This is because, when we consider a graph whose adjacency matrix is given by $\hat{H}$, problem (\ref{eq:kss}) corresponds to finding a subgraph whose connections are ``sparse", i.e., the sum of the edge weights is small.

\subsection{Desirable Matrix $H$}
\label{sec:H}

Before introducing the details of $\hat{H}$, we first show that the next matrix $H$ has the desired property.
Here, let $g$ be a proper distance between distributions that satisfy the following two properties:
\begin{align}
	\begin{split}
	{\text {\rm non-negativity:}} \; & g(p, q) \ge 0 , \\
	{\text {\rm identity of indiscernibles:}} \; & g(p, q) = 0 \Leftrightarrow p = q.
	\end{split}
	\label{eq:g_cond}
\end{align}
Then, we define matrix $H \in \mathbb{R}_+^{D \times D}$ by
\begin{align}
	H_{ij} := \begin{cases}
		g(p_i, q_i) , & (i = j) , \\
		g(p_{ij}, q_{ij}) , & (i \neq j) ,
	\end{cases}
	\label{eq:H}
\end{align}
where $p_i$ and $q_i$ are univariate distributions of the $i$-th feature on $p$ and $q$, respectively.
Moreover, $p_{ij}$ and $q_{ij}$ are the distributions of a pair of features $(x_i, x_j)$ on $p$ and $q$, respectively.
The next theorem guarantees that, by using matrix $H$, we can derive set $S^*$ by solving problem (\ref{eq:kss}).
\begin{theorem}
	\label{th:solution}
	For a matrix $H$ defined in (\ref{eq:H}), the next relation holds when $|{S^*}^\comp| = k$:
	\begin{align}
		{S^*}^\comp \in \argmin_{S^\comp \subseteq [D]} \sum_{i, j \in S^\comp} H_{ij}, \; {\rm s.t.} \; |S^\comp| = k .
	\end{align}
\end{theorem}
All the proofs in this paper can be found in Appendix~\ref{sec:proof}.
We note that, in definition (\ref{eq:H}), we assume that the true distributions $p$ and $q$ are known.
In practice, we do not know these distributions; therefore the desirable matrix $H$ is not accessible.

\subsection{The KS-Matrix}
\label{sec:HKS}

We now turn to designing matrix $\hat{H}$ by utilizing the desirable property of the matrix $H$ defined in (\ref{eq:H}).
Specifically, we answer two questions: what distance function $g$ to use and how we approximate the distance using a limited number of observations $\mathcal{P}$ and $\mathcal{Q}$.
In this study, we propose using the KS statistic as the distance function $g$.
However, we note that, in general, the KS statistic is defined only on distributions over one dimensional real-valued random variables.
Hence, it is not directly applicable to our study because we are interested in the distance between the two dimensional distributions $g(p_{ij}, q_{ij})$.
We resolve this problem by extending the KS statistic to the two-dimensional case.

\subsubsection{Modified KS Statistic in Two Dimensions}

We propose a modified KS statistic that measures a distance between two dimensional distributions $g(p_{ij}, q_{ij})$.
Although there have been some attempts to extend the KS statistic to more than one dimension~\citep{peacock1983two,fasano1987multidimensional,justel1997multivariate,lopes2007two}, they tend to be computationally demanding.
By contrast, the proposed modified statistic can be approximated in $O(L N \log N)$ time where $L$ is an algorithm parameter.
Specifically, we consider projecting the two dimensional feature $(x_i, x_j)$ to one dimension as $r_{ij, \theta} = x_i \cos \theta + x_j \sin \theta$ where $\theta \in [0, \pi]$.
Here, we denote the distributions of $r_{ij, \theta}$ under $p$ and $q$ by $p_{ij, \theta}$ and $q_{ij, \theta}$, respectively.
We then measure the KS statistic between the distributions ${\rm KS}(p_{ij, \theta}, q_{ij, \theta})$.
Because ${\rm KS}(p_{ij, \theta}, q_{ij, \theta})$ depends on the newly introduced parameter $\theta$, we define distance $g$ as the expectation of the KS statistic over parameter $\theta$ assuming that $\theta$ is uniformly random over $[0, \pi]$.
We then define the modified KS statistic as
\begin{align}
	g(p_{ij}, q_{ij}) := \mathbb{E}_{\theta \sim \mathcal{U}(0, \pi)}\left[ {\rm KS}(p_{ij, \theta}, q_{ij, \theta}) \right] .
	\label{eq:g}
\end{align}
We note that this distance $g$ satisfies condition (\ref{eq:g_cond}) because the KS statistic is an $L_\infty$-distance between the distributions.
Because the expectation is a linear operator, it preserves the original property of the $L_\infty$-distance.

\subsubsection{Approximating The Modified KS Statistic}

The exact computation of the modified KS statistic given by (\ref{eq:g}) is difficult because the expectation over $\theta$ is intractable.
Here, we propose approximating the statistic using sampling.
Specifically, we randomly sample $\{\theta_\ell\}_{\ell=1}^L$ from $\mathcal{U}(0, \pi)$ and compute the following as an approximation of $g$:
\begin{align}
	\hat{g}_L(p_{ij}, q_{ij}) := \frac{1}{L} \sum_{\ell=1}^L {\rm KS}(p_{ij, \theta_\ell}, q_{ij, \theta_\ell}).
	\label{eq:g_approx}
\end{align}
Because the one-dimensional KS statistic can be computed in $O(N \log N)$ time, the computation of approximation (\ref{eq:g_approx}) takes only $O(L N \log N)$ time.

The next theorem shows that approximation (\ref{eq:g_approx}) becomes exponentially tight as the number of samplings $L$ increases.
\begin{theorem}
	\label{th:hoeffding}
	For any $\delta > 0$, the next inequality holds:
	\begin{align}
		{\rm Pr}\left( \left| g(p_{ij}, q_{ij}) - \hat{g}_L(p_{ij}, q_{ij}) \right| > \delta \right) \le 2 \exp \left( - 2 \delta^2 L \right) .
	\end{align}
\end{theorem}

\subsubsection{The KS-Matrix}

Using the modified KS statistic, we define a KS-matrix $H \in \mathbb{R}_+^{D \times D}$ as
\begin{align}
	H_{ij} := 
	\begin{cases}
		{\rm KS}(p_i, q_i) , & (i = j) , \\
		\mathbb{E}_{\theta \sim \mathcal{U}(0, \pi)}\left[ {\rm KS}(p_{ij, \theta}, q_{ij, \theta}) \right] , & (i \neq j) . \\
	\end{cases}
	\label{eq:HKS}
\end{align}
We also define an empirical KS-matrix $\hat{H} \in \mathbb{R}_+^{D \times D}$ as
\begin{align}
	\hat{H}_{ij} := 
	\begin{cases}
		{\rm KS}(\hat{p}_i, \hat{q}_i) , & (i = j) , \\
		\frac{1}{L} \sum_{\ell=1}^L {\rm KS}(\hat{p}_{ij, \theta_\ell}, \hat{q}_{ij, \theta_\ell}) , & (i \neq j) . \\
	\end{cases}
	\label{eq:HKShat}
\end{align}
Because the empirical KS-matrix is composed of $O(D^2)$ entires, the overall computation of the empirical KS-matrix $\hat{H}$ takes $O(D^2 L N \log N)$ time.
We note that, because the computation of each matrix entry can be conducted independently, the computation of the matrix can be parallelized easily.

An important property of the empirical KS-matrix in (\ref{eq:HKShat}) is that the solution to problem (\ref{eq:kss}) is identical to ${S^*}^\comp$ under an appropriate condition.
Formally, the next theorem guarantees that $\hat{S} = S^*$ holds when the empirical KS-matrix $\hat{H}$ is sufficiently close to the KS-matrix $H$.
\begin{theorem}{\bf \citep[Theorem 1]{hara2015consistent}}
	\label{th:consistent}
	Let $\eta = \min_{S^\comp \neq {S^*}^{\rm c} : |S^\comp| = k} \sum_{i, j \in S^\comp} H_{ij} - \sum_{i, j \in {S^*}^{\rm c}} H_{ij}$ and assume $\eta > 0$.
	Then, $\hat{S} = S^*$ holds if $\vi H - \hat{H} \vi_\infty \le \eta / 2 k^2$, where $\vi \cdot \vi_\infty$ denotes an element-wise infinity norm of a matrix $\vi M \vi_\infty = \max_{i, j} |M_{ij}|$.
\end{theorem}
We note that the positivity assumption of $\eta$ relates to the uniqueness of feature set $S^*$.
If the assumption is violated, i.e., $\eta = 0$, there exists another feature set $S^{**} \neq S^*$ that attains the same minimum as that of $S^*$, i.e., $\sum_{i, j \in {S^*}^\comp} H_{ij} = \sum_{i', j' \in {S^{**}}^\comp} H_{i'j'}$.
The positivity of $\eta$ assures that $S^*$ is uniquely identifiable.
In Section~\ref{sec:consistent}, we show that $\hat{S} = S^*$ holds for $\eta > 0$ with high probability when the number of observations $N$ and $M$ and the number of samplings $L$ are sufficiently large.
We also discuss when the assumption $\eta > 0$ holds.

\section{Solution Algorithms}
\label{sec:algorithms}

For the proposed empirical KS-matrix $\hat{H}$, Theorem~\ref{th:consistent} guarantees that we can identify feature set $S^*$ by solving problem (\ref{eq:kss}).
The challenge is that the sparsest $k$-subgraph problem in (\ref{eq:kss}) is NP-hard in general~\citep{watrigant2016approx}.
We first review the exact solution method using integer programming, and we then propose a greedy method as a computationally efficient approximation.

\subsection{The Exact Method}

A naive way to solve problem (\ref{eq:kss}) is to use general combinatorial methods.
For instance, we can use an exact method to solve the problem.
The solution can be derived by solving the binary quadratic problem:
\begin{align}
	\hat{\bm{s}} := \argmin_{\bm{s} \in \{0, 1\}^D} \bm{s}^\top \hat{H} \bm{s}, \; {\rm s.t.} \; \bm{1}_D^\top \bm{s} = k .
	\label{eq:ip}
\end{align}
Set $\hat{S}$ can be recovered from the solution by $\hat{S} := \{d \mid \hat{s}_d = 0\}$.
We note that problem (\ref{eq:ip}) is NP-hard in general.
The solution can be derived using state-of-the-art solvers such as the IBM ILOG CPLEX although it may take exponential time.

\subsection{The Greedy Method}

In practice, we can use the greedy method shown in Algorithm~\ref{alg:greedy} to derive a pragmatic solution in polynomial time, as shown in our previous study~\citep{hara2015consistent}.
The advantage of the greedy method is that it runs in only $O((D - k) D)$ time using book keeping.
Let $f(S) := \sum_{i, j \in S^\comp} \hat{H}_{ij}$.
In book keeping, we maintain $\bm{a} \in \R^D$ such that $a_d := \sum_{i \in \tilde{S}^\comp} \hat{H}_{d i}$ for every $d \in \tilde{S}^\comp$.
Then, in every iteration, the value of $f(\tilde{S} \cup \{d\})$ can be computed as $f(\tilde{S} \cup \{d\}) = f(\tilde{S}) - 2a_d + \tilde{H}_{dd}$ which is $O(1)$ time for every $d \in \tilde{S}^\comp$.
Thus, the $\argmin$ operation can be computed in $O(D)$ time.
We then update $\bm{a}$ by $a_{d'} \leftarrow a_{d'} - \hat{H}_{dd'}$ when an update $\tilde{S} \leftarrow \tilde{S} \cup \{d\}$ is executed, which is also $O(D)$ time.
Hence, one iteration in Algorithm~\ref{alg:greedy} runs in $O(D)$ time, and the overall time complexity is $O((D - k) D)$.
We note that this time complexity is far smaller than that required to compute the empirical KS-matrix, which takes $O(D^2 L N \log N)$ time.
The computation time for the greedy method is thus negligible in practice.

\begin{figure}[t]
\centering
\begin{algorithm}[H]
	\caption{Greedy Method}
	\label{alg:greedy}
	\begin{algorithmic}
		\REQUIRE Empirical KS-matrix $\hat{H} \in \R_+^{D \times D}$, integer $k$,
		\ENSURE $\tilde{S} \subseteq [D]$
		\STATE Define $f(S) := \sum_{i, j \in S^\comp} \hat{H}_{ij}$
		\STATE Let $\tilde{S} \leftarrow \emptyset$
		\FOR{$i=1$ to $D - k$}
		\STATE $d \leftarrow \argminl_{d' \in \tilde{S}^\comp} f(\tilde{S} \cup \{d'\})$
		\STATE $\tilde{S} \leftarrow \tilde{S} \cup \{d\}$
		\ENDFOR
	\end{algorithmic}
\end{algorithm}
\end{figure}

Another advantage of the greedy method is its guaranteed approximation ratio.
The next theorem states that, by using the greedy method, we can derive a good approximate solution.
The proof follows from the $(1 - 1/e)$-approximability of the monotone submodular function maximization under the cardinality constraint~\citep{nemhauser1978analysis}.
\begin{theorem}
	\label{th:approx}
	Let $f'(S) := \sum_{i, j \in [D]} \hat{H}_{ij} - \sum_{i, j \in S^\comp} \hat{H}_{ij}$.
	Then, the solution $\tilde{S}$ derived by Algorithm~\ref{alg:greedy} satisfies
	\begin{align}
		f'(\tilde{S}) \ge \left(1 - \frac{1}{e}\right) f'(\hat{S}) .
		\label{eq:approx}
	\end{align}
\end{theorem}

One difficulty with the greedy method is that $k$, the size of ${S^*}^\comp$,  is unknown in most cases.
Therefore, we propose the new heuristic algorithm shown in Algorithm~\ref{alg:greedy_score} to avoid specifying $k$.
In this algorithm, we score feature $x_d$ based on the normalized change in function value $f$ when an element $d$ is added to $\tilde{S}$.
If the addition of $d$ to $\tilde{S}$ significantly reduces the function value, we can conjecture that there is a distribution difference in feature $x_d$.
Formally, we estimate set $S^*$ by $\tilde{S}_t := \{d \mid \hat{s}_d > t\}$ by applying a threshold $t$ to the score $\hat{\bm{s}}$ derived from Algorithm~\ref{alg:greedy_score}.
This procedure is more practical than the original greedy method because $k$ does not need to be specified explicitly.
Threshold $t$ can be determined, for instance, by a visual inspection of the score bar chart.
We note that, similarly to the greedy method, the greedy scoring method runs in $O(D^2)$ time using the same book keeping technique.

Other than greedy methods, one can also use a convex relaxation method~\citep{hara2015consistent} to solve problem~(\ref{eq:kss}).
One can then derive a sparse solution that does not require specifying a threshold at the cost of computation time.

\begin{figure}[t]
\begin{algorithm}[H]
	\caption{Greedy Scoring Method}
	\label{alg:greedy_score}
	\begin{algorithmic}
		\REQUIRE Empirical KS-matrix $\hat{H} \in \R_+^{D \times D}$
		\ENSURE Score vector $\tilde{\bm{s}} \in \R^D$
		\STATE Define $f(S) := \sum_{i, j \in S^\comp} \hat{H}_{ij}$
		\STATE Let $\tilde{S} \leftarrow \emptyset$, $\tilde{\bm{s}} \leftarrow \bm{0}_D$
		\FOR{$i=1$ to $D$}
		\STATE $d \leftarrow \argminl_{d' \in \tilde{S}^\comp} f(\tilde{S} \cup \{d'\})$
		\STATE $\tilde{s}_d \leftarrow (f(\tilde{S}) - f(\tilde{S} \cup \{d\})) \, / \, (D-i+1)$
		\STATE $\tilde{S} \leftarrow \tilde{S} \cup \{d\}$
		\ENDFOR
	\end{algorithmic}
\end{algorithm}
\end{figure}

\section{Theoretical Analysis}
\label{sec:consistent}

We give a feature selection consistency theorem for the estimated set $\hat{S}$ derived by solving problem (\ref{eq:kss}).
Specifically, we show that, under appropriate conditions, the probability of the misspecification $\hat{S} \neq S^*$ decays exponentially as the number of samples $N$ and $M$ and the number of samplings $L$ increase.
In this section, we assume $N = M$ for the ease of discussion.
The results in this section can be naturally extended to the general $N \neq M$ case by replacing $N$ with $\min\{N, M\}$.

The next lemmas show that both the diagonal and off-diagonal elements of the empirical KS-matrix converge to the KS-matrix as the number of samples $N$ and $M$ and the number of samplings $L$ increase.
\begin{lemma}[Convergence of diagonal elements]
	\label{lem:diag}
	Assume that $N = M$.
	The following inequality then holds for any $\delta > 0$:
	\begin{align}
		{\rm Pr} \left( \left| {\rm KS}(p_i, q_i) - {\rm KS}(\hat{p}_i, \hat{q}_i) \right| > \delta \right) \le 4 \exp \left( - \frac{\delta^2}{2} N \right) .
	\end{align}
\end{lemma}

\begin{lemma}[Convergence of off-diagonal elements]
	\label{lem:offdiag}
	Assume that $N = M$.
	There exists $A_{ij}, B_{ij} > 0$ such that the following inequality holds for any $\delta > 0$:
	\begin{align}
		{\rm Pr}\left( \left| g(p_{ij}, q_{ij}) - \hat{g}_L(\hat{p}_{ij}, \hat{q}_{ij}) \right| > \delta \right) \le 2 \exp \left( - 2 C_{ij, \delta} N \right) + 2 \exp \left( - \frac{\delta^2}{2} L \right) ,
	\end{align}
	where $C_{ij, \delta} := \left(\sqrt{A_{ij}^2 + 2 B_{ij} \delta} - A_{ij} \right)^2 / 2 B_{ij}^2$.
\end{lemma}

The probability of misspecification $\hat{S} \neq S^*$ then follows by combining Lemma~\ref{lem:diag} and \ref{lem:offdiag} with Theorem~\ref{th:consistent}.
\begin{theorem}[Consistency of $\hat{S}$]
	\label{th:main}
	Assume that $N = M$.
	Let $\eta$ be the parameter defined in Theorem~\ref{th:consistent} and assume $\eta > 0$.
	Then, the probability of misspecification $\hat{S} \neq S^*$ is bounded as
	\begin{align}
		{\rm Pr}(\hat{S} \neq S^*) \le 4 D \exp \left( - \frac{\eta^2}{8 k^4} N \right) + D (D - 1) \left\{ \exp \left( - 2 C_{\eta / 2 k^2} N \right) + \exp \left( - \frac{\eta^2}{8 k^4} L \right) \right\} ,
		\label{eq:main}
	\end{align}
	where $C_{\eta / 2 k^2} := \min_{i, j \in [D] : i > j} C_{ij, \eta / 2 k^2}$.
\end{theorem}

Theorem~\ref{th:main} indicates that the probability of misspecification $\hat{S} \neq S^*$ decays exponentially as the number of samples $N$ and $M$ and the number of samplings $L$ increase.
This bound gives us a guideline as to how many samples $N$ and $M$ as well as samplings $L$ are required to maintain the misspecification probability within a desired level.
\begin{corollary}
	\label{cor:n}
	Assume that $N = M$.
	To guarantee ${\rm Pr}(\hat{S} \neq S^*) \le \epsilon$ for $\epsilon > 0$, we require
	\begin{align}
		N = O\left(\max\left\{\frac{k^4}{\eta^2}, \frac{1}{C_{\eta/2k^2}}\right\} \log \frac{D}{\epsilon} \right), \qquad L = O\left(\frac{k^4}{\eta^2} \log \frac{D}{\epsilon} \right).
	\end{align}
\end{corollary}
Here, we note that $1 / C_{\eta / 2k^2} = B^2 / ( \sqrt{A^2 + B \eta / k^2} - A )^2 \approx B k^2 / \eta$ holds for some $A, B > 0$.
Hence, the order of $N$ can be approximated as $N \approx O\left(\max\left\{k^4/\eta^2, k^2/\eta\right\} \log D / \epsilon \right)$.

One key assumption in Theorem~\ref{th:main} is the strict positivity of $\eta$, which assures the uniqueness of $S^*$.
The next theorems give the necessary and sufficient conditions for $\eta > 0$.
\begin{theorem}[Necessary condition]
	\label{th:consistency_necc}
	If $\eta > 0$, one of $({\rm N1})$ and $({\rm N2})$ holds for any $d \in S^*$:
	\begin{align}
		({\rm N1}) \; H_{dd} > 0 , \qquad ({\rm N2}) \; \exists d' \in [D] \setminus \{d\} , \; H_{d d'} > 0 .
	\end{align}
\end{theorem}

\begin{theorem}[Sufficient condition]
	\label{th:consistency_suff}
	$\eta > 0$ holds when one of $({\rm S1})$ and $({\rm S2})$ holds for any $d \in S^*$:
	\begin{align} 
		({\rm S1}) \; H_{dd} > 0 , \qquad ({\rm S2}) \; \forall d' \in [D] \! \setminus \{d\} , \; H_{dd'} > 0 .
	\end{align}
\end{theorem}

Conditions (N1) and (N2) require the distribution difference to be observed on each pair of features.
Note that this is not a restrictive assumption in practice.
Conditions (N1) and (N2) are violated only when the difference appears on the distribution of more than two variables, i.e., $p(x_d, x_{d'}, x_{d''}) \neq q(x_d, x_{d'}, x_{d''})$ holds while $p(\bm{x}_T) = q(\bm{x}_T)$ for any $T \subsetneq \{d, d', d''\}$.
Intuitively, these cases are negligible in practice as they require the distributions $p$ and $q$ to have very specific structures.
The following theorem guarantees that this intuition is correct in the Gaussian case.
Indeed, conditions (N1) and (N2) hold for any distribution differences under Problem~\ref{prob:anomaly} with a Gaussian distribution.
\begin{theorem}
	\label{th:consistency_gauss}
	When both $p$ and $q$ are Gaussian, one of (N1) and (N2) holds for any $d \in S^*$.
\end{theorem}

\section{Relation to Current Methods}
\label{sec:relation}

The proposed method can be interpreted as a generalization of our previous method~\citep{hara2015consistent}, which is the first algorithm that uses the sparsest $k$-subgraph problem for different-feature selection.
Unlike the proposed method, the previous method has limited applicability due to the Gaussian assumption.
In the previous method, we assumed Gaussian distributions on $p$ and $q$, and defined matrix $\hat{H} \in \mathbb{R}_+^{D \times D}$ by $\hat{H}_{ij} := |C_{ij}^\mathcal{P} - C_{ij}^\mathcal{Q}|$, where matrices $C^\mathcal{P}$ and $C^\mathcal{Q}$ are the covariance or precision matrices of datasets $\mathcal{P}$ and $\mathcal{Q}$, respectively.
This corresponds to using an approximation of the KL-divergence as the measurement of the difference between the two distributions rather than the KS statistic.
Indeed, $\hat{H}_{ij}$ defined above corresponds to the lower bound of the KL-divergence between the two Gaussian distributions under the specific case described in the next proposition.
\begin{proposition}
	\label{prop:rel}
	Suppose $p$ and $q$ are Gaussian distributions with the same mean $\bm{\mu} \in \R^D$: $p(\bm{x}) := \mathcal{N}(\bm{\mu}, \Sigma)$ and $q(\bm{x}) := \mathcal{N}(\bm{\mu}, \Gamma)$. When both $\Sigma$ and $\Gamma$ are invertible and have diagonal components equal to one, $|\Sigma_{ij} - \Gamma_{ij}|$ is a lower bound of the KL-divergence ${\rm KL}[p_{ij} || q_{ij}]$ up to a constant term.
\end{proposition}

\section{Experiments}
\label{sec:simu}

We evaluated the different-feature selection performance of the proposed method with respect to both its accuracy and runtime.
We first give illustrative examples with synthetic data that describe the advantages and disadvantages of the proposed method.
We then present experimental results on UCI datasets and on a quantum system anomaly detection application.
All experiments were conducted using a 16-core VM with an Intel Xeon E312xx, 16GB of RAM, and Ubuntu 15.04.

\paragraph{Baseline Methods:} 
We compared the proposed method to four baseline methods.
The first three are the Gaussian-based methods MT~\citep{taguchi2000new}, Id\'e'09~\citep{ide2009proximity}, and Hara'15~\citep{hara2015consistent}, and the last one is the nonparametric method SPARDA~\citep{mueller2015principal}.
See Appendix~\ref{sec:baseline} for the details of each method.

\paragraph{Implementations:}
In the experiments, we used the greedy scoring method (Algorithm~\ref{alg:greedy_score}) as the proposed method.
The proposed method and Gaussian-based methods were implemented in Python.
SPARDA was implemented in C++ based on the MATLAB code \texttt{fastSPARDA.m}, which is available on the author's website (\url{http://www.mit.edu/~jonasm/}).
For the proposed method, we set the number of samplings $L = 10$.
For SPARDA, because the relax and tighten procedure was too slow, we used the projected gradient ascent, which runs in $O(D N + N \log N)$ time per iteration.
Because the projected gradient ascent tends to be trapped by local optima, we used five random restarts.
We set the regularization parameter candidate for SPARDA to $\{0, 10^{-4}, 10^{-3}, 10^{-2}, 10^{-1}\}$ and selected the optimal one using five-fold cross validation.

\paragraph{Evaluation Metric:}
Each method outputs a $D$-dimensional score vector that describes how likely it is that the corresponding feature has changed.
We compare the score vector to the ground truth features $S^*$, and then measure the area under the receiver operating characteristic curve (AUROC).
AUROC$=1$ means that the features are correctly identified with high scores.
We note that AUROC does not require specifying the threshold on the score, and hence it is a desirable evaluation metric.

\subsection{Illustrative Examples}
\label{sec:example}

Here, we show the advantages and disadvantages of the proposed method on synthetic experiments.
We also present a runtime comparison of the proposed method and SPARDA.

\paragraph{[Example 1] Gaussian with Covariance Change:}
In the first example, we used Gaussian data.
We generated synthetic data as follows:
Let $\Theta$ be a $20 \times 20$ randomly generated matrix from $\mathcal{U}(-1, 1)$.
We then computed $\Sigma = \Theta^\top \Theta$ and normalized the diagonal of $\Sigma$ to be one.
Furthermore, we generated $20$-dimensional data from the distributions $p(\bm{x}) = \mathcal{N}(\bm{0}_{20}, \Sigma)$ and $q(\bm{x}) = \mathcal{N}(\bm{0}_{20}, \Sigma')$, where $\Sigma'_{11} = 0.49 \Sigma_{11} + 0.09 \Sigma_{22} + 0.21 \Sigma_{12}$, $\Sigma'_{1d} = 0.7 \Sigma_{1d} + 0.3 \Sigma_{dd}$ for $d \in [20] \setminus \{1\}$, and $\Sigma_{dd'} = \Sigma'_{dd'}$ otherwise.
In this setting, $S^* = \{1\}$ is the solution to Problem~\ref{prob:anomaly}.
We set the numbers of data points in $\mathcal{P}$ and $\mathcal{Q}$ to be equal, i.e., $N = M$.
Then, we randomly generated datasets 100 times for several different dataset sizes $N$.

\figurename~\ref{fig:example_auroc1} shows the average AUROC of each method over 100 random data realizations.
Id\'e'09 and Hara'15 converged to an average AUROC = 1 around $N=10^2$ and $N = 10^3$, respectively.
The proposed method attained an average AUROC = 1 around $N = 3 \times 10^3$, which is slower than the previous two methods.
This shows that the use of the correct parametric model is advantageous in different-feature selection.
However, we note that the proposed method provided a consistent result with large sample sizes, as implied by Theorem~\ref{th:main}.
In other words, the proposed method can be an alternative to Gaussian-based methods when there is a sufficiently large number of samples.
Note that SPARDA attained a comparable but a slightly lower average AUROC.

\paragraph{[Example 2] Gaussian Mixture with Mixture Rate Change:}
In the second example, we used non-Gaussian data to show the advantages of the proposed method.
In this example, we generated $20$-dimensional data from the Gaussian mixture distributions $p$ and $q$ with different mixture rates for feature $x_1$.
Let $p(\bm{u}) = \mathcal{N}(\bm{0}_{20}, \Sigma)$ be a $20$-dimensional Gaussian distribution.
We defined $p(x_d | u_d) = 0.5 \delta(x_d - u_d/3 - 4/3) + 0.5 \delta(x_d - u_d/3 + 4/3)$ for $d=1$ and $p(x_d | u_d) = \delta(x_d - u_d)$ otherwise, where $\delta(\cdot)$ is a delta function.
We also defined $q(x_d | u_d) = 0.35 \delta(x_d - u_d/3 - 4/3) + 0.35 \delta(x_d - u_d/3 + 4/3) + 0.3 \delta(x_d - u_d/3)$ for $d = 1$ and $q(x_d | u_d) = \delta(x_d - u_d)$ otherwise.
In this setting, $S^* = \{1\}$ is the solution to Problem~\ref{prob:anomaly}.
Note that the change from $p$ to $q$ causes variance change in feature $x_1$; therefore, it can be detected using the Gaussian-based methods.

\figurename~\ref{fig:example_auroc2} shows the advantage of the proposed method.
It attained an average AUROC = 1 around $N=5 \times 10^2$, which is a fast convergence compared to the Gaussian-based methods.
Id\'e'09 required $N=10^3$ to attain an average AUROC = 1, and MT and Hara'15 required more samples.
This indicates that the proposed method can detect the complex distribution difference effectively due to its nonparametric nature.
Thus, it performed well with non-Gaussian data where the Gaussian-based methods performed poorly.
Note that the performance of SPARDA was worse than the proposed method for small sample sizes, whereas its average AUROC converged to one for large sample sizes.

\begin{figure}[t]
	\centering
	\subfigure[Example1: Gaussian w/ Covariance Change]{
	\begin{tikzpicture}
	\begin{semilogxaxis}[
		scale=1.0,
		xlabel={\# of samples $N$},
		xmin=10,
		xmax=10000,
		ylabel style={align=center},
		ylabel={Average AUROC},
		width=0.45\textwidth,
		height=4.5cm,
		legend style={at={(1.00,1.35)}, font=\fontsize{7}{5}\selectfont},
		legend columns=3,
	]
	\addplot [color=blue, line width=1.2pt, mark=*, mark options={solid, scale=1.0, fill=blue}, solid]
	table [x=num,y=auc,meta=label,col sep=comma,skip coords between index={13}{65}]{cov_p03_result.csv};
	\addplot [color=cyan, line width=1.2pt, mark=pentagon*, mark options={solid, scale=1.0, fill=cyan}, dash pattern=on 1pt off 3pt on 3pt off 3pt]
	table [x=num,y=auc,meta=label,col sep=comma,skip coords between index={0}{13},skip coords between index={26}{65}]{cov_p03_result.csv};
	\addplot [color=magenta, line width=1.2pt, mark=triangle*, mark options={solid, scale=1.0, fill=magenta}, dotted]
	table [x=num,y=auc,meta=label,col sep=comma,skip coords between index={0}{26},skip coords between index={39}{65}]{cov_p03_result.csv};
	\addplot [color=green, line width=1.2pt, mark=diamond*, mark options={solid, scale=1.0, fill=green}, dash pattern=on 2pt off 4pt on 4pt off 4pt]
	table [x=num,y=auc,meta=label,col sep=comma,skip coords between index={0}{39},skip coords between index={52}{65}]{cov_p03_result.csv};
	\addplot [color=red, line width=1.2pt, mark=square*, mark options={solid, scale=1.0, fill=red}, dashed]
	table [x=num,y=auc,meta=label,col sep=comma,skip coords between index={0}{52}]{cov_p03_result.csv};
	\legend{Proposed, MT, Ide'09, Hara'15, SPARDA}
	\end{semilogxaxis}
	\end{tikzpicture}
	\label{fig:example_auroc1}
	}
	\hfill
	\subfigure[Example2: Gaussian Mixture w/ Rate Change]{
	\begin{tikzpicture}
	\begin{semilogxaxis}[
		scale=1.0,
		xlabel={\# of samples $N$},
		xmin=10,
		xmax=10000,
		ylabel style={align=center},
		ylabel={Average AUROC},
		width=0.45\textwidth,
		height=4.5cm,
		legend style={at={(1.00,1.35)}, font=\fontsize{7}{5}\selectfont},
		legend columns=3,
	]
	\addplot [color=blue, line width=1.2pt, mark=*, mark options={solid, scale=1.0, fill=blue}, solid]
	table [x=num,y=auc,meta=label,col sep=comma,skip coords between index={13}{65}]{mix_p03_result.csv};
	\addplot [color=cyan, line width=1.2pt, mark=pentagon*, mark options={solid, scale=1.0, fill=cyan}, dash pattern=on 1pt off 3pt on 3pt off 3pt]
	table [x=num,y=auc,meta=label,col sep=comma,skip coords between index={0}{13},skip coords between index={26}{65}]{mix_p03_result.csv};
	\addplot [color=magenta, line width=1.2pt, mark=triangle*, mark options={solid, scale=1.0, fill=magenta}, dotted]
	table [x=num,y=auc,meta=label,col sep=comma,skip coords between index={0}{26},skip coords between index={39}{65}]{mix_p03_result.csv};
	\addplot [color=green, line width=1.2pt, mark=diamond*, mark options={solid, scale=1.0, fill=green}, dash pattern=on 2pt off 4pt on 4pt off 4pt]
	table [x=num,y=auc,meta=label,col sep=comma,skip coords between index={0}{39},skip coords between index={52}{65}]{mix_p03_result.csv};
	\addplot [color=red, line width=1.2pt, mark=square*, mark options={solid, scale=1.0, fill=red}, dashed]
	table [x=num,y=auc,meta=label,col sep=comma,skip coords between index={0}{52}]{mix_p03_result.csv};
	\legend{Proposed, MT, Ide'09, Hara'15, SPARDA}
	\end{semilogxaxis}
	\end{tikzpicture}
	\label{fig:example_auroc2}
	}
	\caption{Comparison of AUROC on two synthetic datasets}
	\label{fig:example_auroc}
\end{figure}
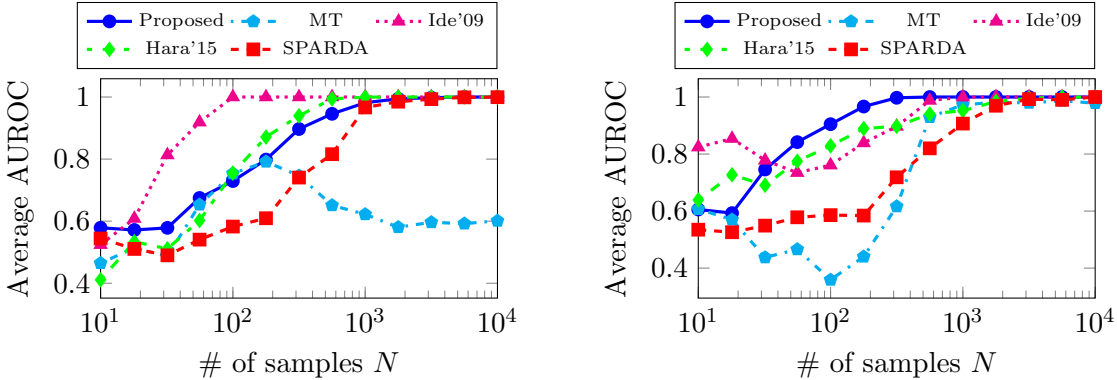

\begin{figure}[t]
	\centering
	\subfigure[Example1: Gaussian w/ Covariance Change]{
	\begin{tikzpicture}
	\begin{axis}[
		scale=1.0,
		xmode=log,
		ymode=log,
		xlabel={\# of samples $N$},
		xmin=10,
		xmax=10000,
		ylabel style={align=center},
		ylabel={Average Runtime (sec)},
		width=0.44\textwidth,
		height=4.5cm,
		legend style={at={(1.05,1.35)}, font=\fontsize{6}{5}\selectfont},
		legend columns=2,
	]
	\addplot [color=blue, line width=1.2pt, mark=*, mark options={solid, scale=1.0, fill=blue}, solid]
	table [x=num,y=time,meta=label,col sep=comma,skip coords between index={0}{13},skip coords between index={26}{52}]{cov_p03_time.csv};
	\addplot [color=blue, line width=1.2pt, mark=*, mark options={solid, scale=1.0, fill=white}, solid]
	table [x=num,y=time,meta=label,col sep=comma,skip coords between index={13}{52}]{cov_p03_time.csv};
	\addplot [color=red, line width=1.2pt, mark=square*, mark options={solid, scale=1.0, fill=red}, dashed]
	table [x=num,y=time,meta=label,col sep=comma,skip coords between index={0}{39}]{cov_p03_time.csv};
	\addplot [color=red, line width=1.2pt, mark=square*, mark options={solid, scale=1.0, fill=white}, dashed]
	table [x=num,y=time,meta=label,col sep=comma,skip coords between index={0}{26},skip coords between index={39}{52}]{cov_p03_time.csv};
	\legend{Proposed (10-parallel), Proposed (single), SPARDA (10-parallel), SPARDA (single)}
	\end{axis}
	\end{tikzpicture}
	\label{fig:example_time1}
	}
	\subfigure[Example2: Gaussian Mixture w/ Rate Change]{
	\begin{tikzpicture}
	\begin{axis}[
		scale=1.0,
		xmode=log,
		ymode=log,
		xlabel={\# of samples $N$},
		xmin=10,
		xmax=10000,
		ylabel style={align=center},
		ylabel={Average Runtime (sec)},
		width=0.44\textwidth,
		height=4.5cm,
		legend style={at={(1.05,1.35)}, font=\fontsize{6}{5}\selectfont},
		legend columns=2,
	]
	\addplot [color=blue, line width=1.2pt, mark=*, mark options={solid, scale=1.0, fill=blue}, solid]
	table [x=num,y=time,meta=label,col sep=comma,skip coords between index={0}{13},skip coords between index={26}{52}]{mix_p03_time.csv};
	\addplot [color=blue, line width=1.2pt, mark=*, mark options={solid, scale=1.0, fill=white}, solid]
	table [x=num,y=time,meta=label,col sep=comma,skip coords between index={13}{52}]{mix_p03_time.csv};
	\addplot [color=red, line width=1.2pt, mark=square*, mark options={solid, scale=1.0, fill=red}, dashed]
	table [x=num,y=time,meta=label,col sep=comma,skip coords between index={0}{39}]{mix_p03_time.csv};
	\addplot [color=red, line width=1.2pt, mark=square*, mark options={solid, scale=1.0, fill=white}, dashed]
	table [x=num,y=time,meta=label,col sep=comma,skip coords between index={0}{26},skip coords between index={39}{52}]{mix_p03_time.csv};
	\legend{Proposed (10-parallel), Proposed (single), SPARDA (10-parallel), SPARDA (single)}
	\end{axis}
	\end{tikzpicture}
	\label{fig:example_time2}
	}
	\caption{Comparison of the runtimes on two synthetic datasets: Runtimes for single-thread and ten-thread implementations were measured.}
	\label{fig:example_runtime}
\end{figure}
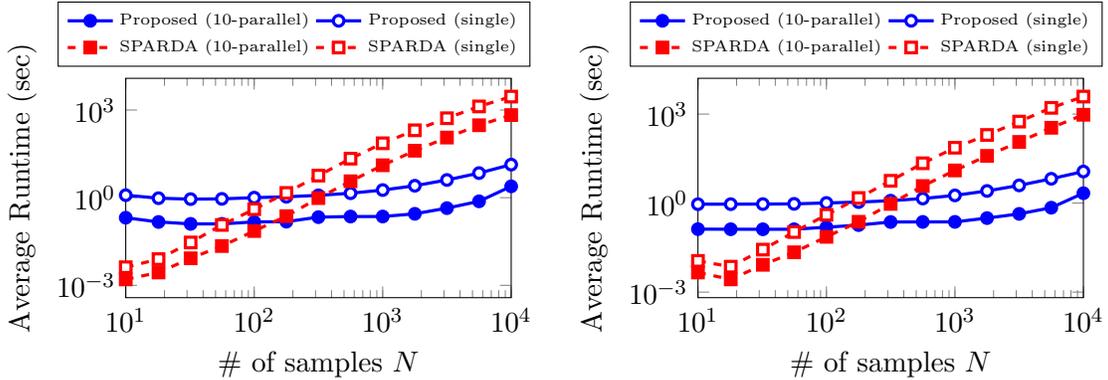

\paragraph{Runtime Comparison:}
\figurename~\ref{fig:example_runtime} shows the entire runtime of the proposed method and SPARDA for two example cases.
For comparison, we used both single-thread and ten-thread implementations.
In the ten-thread implementation, the computation of the empirical KS-matrix $\hat{H}$ was parallelized in the proposed method, whereas the parameter search with cross validation and random restarts were parallelized in SPARDA.

From \figurename~\ref{fig:example_runtime}, we find that the proposed method was significantly faster than SPARDA for large sample sizes.
This was because the proposed method has low time complexity and does not require any extra computation for model selection.
For $N \geq 10^3$, with both the single-thread and ten-thread implementations, the proposed method was more than 100 times faster than SPARDA.
Together with \figurename~\ref{fig:example_auroc}, this result shows that the proposed method could provide consistent solutions in more than 100 times less runtime.
By contrast, SPARDA was computationally advantageous for small sample sizes.

\subsection{Experiments on UCI Datasets}
\label{sec:uci}

Here, we present experimental results on five real-world datasets from the UCI repository~\citep{Lichman:2013}.
The list of datasets is shown in \tablename~\ref{tab:data}.
These datasets are non-Gaussian and are, therefore, appropriate for evaluating the effectiveness of the proposed method.

We constructed the datasets $\mathcal{P}$ and $\mathcal{Q}$ from each dataset, each of which consists of randomly chosen $N=M=1,000$ data points without overlap.
For dataset $\mathcal{Q}$, we randomly selected a feature subset $S^* \subset [D]$ with $|S^*| = 3$ and modified the distribution of $\bm{x}_{S^*}$.
Specifically, for $i \in S^*$ and $j \in {S^*}^\comp$, we applied one of the following five changes:
\begin{enumerate}
	\item[(i)] Mean Shift: $x_i \leftarrow x_i + c$;
	\item[(ii)] Variance Change: $x_i \leftarrow x_i + c \epsilon, \; \epsilon \sim \mathcal{N}(0, 1)$;
	\item[(iii)] Covariance Change: $x_i \leftarrow (1 - c)x_i + c x_j $;
	\item[(iv)] Covariance Change (Conditional): $x_i \leftarrow (1 - c)x_i + c x_j$ when $x_j \leq v$;
	\item[(v)] Covariance Change (No Variance Change): $x_i \leftarrow w (1 - c)x_i + w c x_j $.
\end{enumerate}
Here, $c \in [0, 1]$ is the parameter that controls the difference level, $v$ is the $25\%$ quantile of $x_j$ in dataset $\mathcal{Q}$, and $w$ is a scalar factor that maintains the variance of $x_i$ unchanged.
Note that these changes affect the mean or covariance of the distribution; thus, they can be detected using the Gaussian-based methods.

\begin{table}[t]
	\centering
	\tblcaption{Datasets from the UCI repository. Here, $D_0$ is the number of features, $N_0$ is the number of data points, and $D$ is the number of effective features after screening; we removed features that had less than 10 different values. Each dataset was normalized so that the mean of each feature is zero and its variance is one. }
	\label{tab:data}
	\centering
	\begin{tabular}{c|rrr}
		\hline
		& $D_0$ & $D$ & $N_0$ \\
		\hline
		CASP & 10 & 10 & 45730\\
		CBM~\citep{Coraddu2013Machine} & 18 & 13 & 11934\\
		Diagnosis & 48 & 48 & 58509\\
		MiniBooNE & 50 & 50 & 130065\\
		Statlog & 37 & 36 & 6435\\
		\hline
	\end{tabular}
\end{table}

\begin{table*}[p]
	\caption{Average AUROC $\pm$ standard deviation on 20 random data realizations from five UCI datasets. Proposed (exact) is a referential result with the exact solution of (\ref{eq:kss}) derived using IBM ILOG CPLEX. The highest AUROC of the five methods is shown in bold. The best results and other results were compared using a t-test (5\%), and results that were not rejected are also highlighted.}
	\label{tab:result_uci1}
	\centering
	\small
	\begin{tabular}{cc|c|ccccc}
		\multicolumn{8}{c}{(i) Mean Shift} \\
		& $c$ & \shortstack{Proposed\\(exact)} & Proposed & MT & Id\'e'09 & Hara'15 & SPARDA \\
		\hline
		\multirow{3}{*}{\rotatebox[origin=c]{90}{CASP}}
& .1 & $.93 \pm .15$ & \bm{$.92 \pm .17}$ & $.60 \pm .16$ & $.51 \pm .24$ & $.48 \pm .24$ & $.63 \pm .19$ \\
& .3 & $1.0 \pm .00$ & \bm{$1.0 \pm .00}$ & $.67 \pm .23$ & $.51 \pm .24$ & $.48 \pm .24$ & $.86 \pm .20$ \\
& .5 & $1.0 \pm .00$ & \bm{$1.0 \pm .00}$ & $.72 \pm .21$ & $.51 \pm .24$ & $.48 \pm .24$ & \bm{$.96 \pm .08}$ \\
		\hline
		\multirow{3}{*}{\rotatebox[origin=c]{90}{CBM}}
& .1 & $.98 \pm .07$ & \bm{$.91 \pm .13}$ & $.78 \pm .16$ & $.50 \pm .19$ & $.49 \pm .21$ & $.66 \pm .33$ \\
& .3 & $1.0 \pm .00$ & \bm{$1.0 \pm .00}$ & $.89 \pm .16$ & $.50 \pm .19$ & $.49 \pm .21$ & \bm{$.93 \pm .19}$ \\
& .5 & $1.0 \pm .00$ & \bm{$1.0 \pm .00}$ & \bm{$.92 \pm .17}$ & $.50 \pm .19$ & $.49 \pm .22$ & \bm{$1.0 \pm .00}$ \\
        \hline
        \multirow{3}{*}{\rotatebox[origin=c]{90}{\shortstack{Diag\\nosis}}}
& .1 & $.96 \pm .08$ & \bm{$1.0 \pm .00}$ & $.61 \pm .24$ & $.54 \pm .15$ & $.50 \pm .18$ & $.49 \pm .18$ \\
& .3 & $1.0 \pm .00$ & \bm{$1.0 \pm .00}$ & $.63 \pm .26$ & $.54 \pm .15$ & $.50 \pm .18$ & $.60 \pm .26$ \\
& .5 & $1.0 \pm .00$ & \bm{$1.0 \pm .00}$ & $.71 \pm .23$ & $.54 \pm .15$ & $.50 \pm .18$ & $.68 \pm .29$ \\
        \hline
        \multirow{3}{*}{\rotatebox[origin=c]{90}{\shortstack{Mini\\BooNE}}}
& .1 & $1.0 \pm .00$ & \bm{$1.0 \pm .00}$ & $.72 \pm .33$ & $.55 \pm .15$ & $.43 \pm .16$ & $.60 \pm .32$ \\
& .3 & $1.0 \pm .00$ & \bm{$1.0 \pm .00}$ & $.75 \pm .35$ & $.55 \pm .15$ & $.43 \pm .16$ & $.73 \pm .32$ \\
& .5 & $1.0 \pm .00$ & \bm{$1.0 \pm .00}$ & $.80 \pm .31$ & $.55 \pm .15$ & $.43 \pm .16$ & $.74 \pm .36$ \\
        \hline
        \multirow{3}{*}{\rotatebox[origin=c]{90}{\shortstack{Stat\\log}}}
& .1 & $1.0 \pm .00$ & \bm{$1.0 \pm .00}$ & $.85 \pm .11$ & $.52 \pm .13$ & $.52 \pm .17$ & $.66 \pm .27$ \\
& .3 & $1.0 \pm .00$ & \bm{$1.0 \pm .00}$ & $.93 \pm .03$ & $.52 \pm .13$ & $.52 \pm .17$ & \bm{$.91 \pm .24}$ \\
& .5 & $1.0 \pm .00$ & \bm{$1.0 \pm .00}$ & $.96 \pm .01$ & $.52 \pm .13$ & $.52 \pm .17$ & \bm{$1.0 \pm .00}$ \\
        \hline
        \\
		\multicolumn{8}{c}{(ii) Variance Change} \\
		& $c$ & \shortstack{Proposed\\(exact)} & Proposed & MT & Id\'e'09 & Hara'15 & SPARDA \\
		\hline
		\multirow{3}{*}{\rotatebox[origin=c]{90}{CASP}}
& .1 & $.49 \pm .17$ & \bm{$.50 \pm .26}$ & \bm{$.64 \pm .15}$ & $.52 \pm .23$ & \bm{$.57 \pm .23}$ & \bm{$.62 \pm .17}$ \\
& .3 & $.89 \pm .14$ & \bm{$.93 \pm .11}$ & $.84 \pm .15$ & $.54 \pm .24$ & $.82 \pm .13$ & $.67 \pm .21$ \\
& .5 & $.99 \pm .05$ & \bm{$.98 \pm .07}$ & $.88 \pm .14$ & $.55 \pm .23$ & $.93 \pm .07$ & $.70 \pm .22$ \\
		\hline
		\multirow{3}{*}{\rotatebox[origin=c]{90}{CBM}}
& .1 & $.92 \pm .10$ & \bm{$.92 \pm .12}$ & \bm{$.85 \pm .16}$ & $.51 \pm .19$ & \bm{$.85 \pm .07}$ & $.30 \pm .15$ \\
& .3 & $.96 \pm .09$ & \bm{$.91 \pm .13}$ & \bm{$.94 \pm .11}$ & $.57 \pm .21$ & \bm{$.94 \pm .07}$ & $.26 \pm .18$ \\
& .5 & $.98 \pm .07$ & \bm{$.93 \pm .12}$ & \bm{$.98 \pm .05}$ & $.62 \pm .19$ & \bm{$.99 \pm .03}$ & $.62 \pm .33$ \\
        \hline
        \multirow{3}{*}{\rotatebox[origin=c]{90}{\shortstack{Diag\\nosis}}}
& .1 & $.54 \pm .13$ & \bm{$.63 \pm .15}$ & \bm{$.60 \pm .25}$ & \bm{$.57 \pm .20}$ & $.52 \pm .19$ & $.41 \pm .11$ \\
& .3 & $.80 \pm .15$ & \bm{$.90 \pm .12}$ & $.65 \pm .28$ & $.62 \pm .22$ & $.61 \pm .16$ & $.47 \pm .12$ \\
& .5 & $.96 \pm .08$ & \bm{$1.0 \pm .00}$ & $.71 \pm .26$ & $.63 \pm .25$ & $.70 \pm .14$ & $.48 \pm .14$ \\
        \hline
        \multirow{3}{*}{\rotatebox[origin=c]{90}{\shortstack{Mini\\BooNE}}}
& .1 & $.96 \pm .07$ & \bm{$.95 \pm .10}$ & $.71 \pm .34$ & $.63 \pm .23$ & $.80 \pm .13$ & $.42 \pm .23$ \\
& .3 & $1.0 \pm .00$ & \bm{$1.0 \pm .00}$ & $.76 \pm .34$ & $.71 \pm .28$ & $.86 \pm .16$ & $.34 \pm .23$ \\
& .5 & $1.0 \pm .00$ & \bm{$1.0 \pm .00}$ & $.80 \pm .32$ & $.73 \pm .29$ & \bm{$.92 \pm .17}$ & $.35 \pm .27$ \\
        \hline
        \multirow{3}{*}{\rotatebox[origin=c]{90}{\shortstack{Stat\\log}}}
& .1 & $.60 \pm .16$ & $.76 \pm .16$ & \bm{$.87 \pm .09}$ & $.60 \pm .13$ & $.55 \pm .18$ & $.58 \pm .20$ \\
& .3 & $.90 \pm .16$ & \bm{$.97 \pm .07}$ & $.94 \pm .03$ & $.92 \pm .05$ & $.89 \pm .12$ & $.56 \pm .21$ \\
& .5 & $1.0 \pm .00$ & \bm{$1.0 \pm .00}$ & $.96 \pm .01$ & \bm{$1.0 \pm .00}$ & \bm{$1.0 \pm .00}$ & $.63 \pm .24$ \\
        \hline
	\end{tabular}
\end{table*}

\begin{table*}[p]
	\caption{Average AUROC $\pm$ standard deviation on 20 random data realizations from five UCI datasets. Proposed (exact) is a referential result with the exact solution of (\ref{eq:kss}) derived using IBM ILOG CPLEX. The highest AUROC of the five methods is shown in bold. The best results and other results were compared using a t-test (5\%), and results that were not rejected are also highlighted.}
	\label{tab:result_uci2}
	\centering
	\small
	\begin{tabular}{cc|c|ccccc}
		\multicolumn{8}{c}{(iii) Covariance Change} \\
		& $c$ & \shortstack{Proposed\\(exact)} & Proposed & MT & Id\'e'09 & Hara'15 & SPARDA \\
		\hline
		\multirow{3}{*}{\rotatebox[origin=c]{90}{CASP}}
& .1 & $.69 \pm .17$ & \bm{$.80 \pm .15}$ & $.58 \pm .16$ & $.53 \pm .24$ & $.63 \pm .23$ & $.63 \pm .18$ \\
& .3 & $.90 \pm .12$ & \bm{$.95 \pm .07}$ & $.68 \pm .16$ & $.57 \pm .25$ & $.85 \pm .14$ & $.75 \pm .18$ \\
& .5 & $.96 \pm .09$ & \bm{$.98 \pm .05}$ & $.66 \pm .20$ & $.61 \pm .26$ & $.92 \pm .11$ & $.77 \pm .19$ \\
		\hline
		\multirow{3}{*}{\rotatebox[origin=c]{90}{CBM}}
& .1 & $.91 \pm .11$ & \bm{$.92 \pm .10}$ & $.50 \pm .26$ & $.50 \pm .19$ & $.69 \pm .14$ & $.48 \pm .11$ \\
& .3 & $1.0 \pm .00$ & \bm{$1.0 \pm .00}$ & $.71 \pm .16$ & $.52 \pm .20$ & $.81 \pm .12$ & $.62 \pm .15$ \\
& .5 & $.96 \pm .09$ & \bm{$.97 \pm .07}$ & $.74 \pm .20$ & $.53 \pm .22$ & $.83 \pm .12$ & $.70 \pm .13$ \\
        \hline
        \multirow{3}{*}{\rotatebox[origin=c]{90}{\shortstack{Diag\\nosis}}}
& .1 & $.76 \pm .15$ & \bm{$.88 \pm .11}$ & $.67 \pm .20$ & $.63 \pm .15$ & $.58 \pm .19$ & $.41 \pm .14$ \\
& .3 & $.96 \pm .08$ & \bm{$.98 \pm .05}$ & $.65 \pm .26$ & $.69 \pm .17$ & $.79 \pm .13$ & $.47 \pm .14$ \\
& .5 & $.97 \pm .06$ & \bm{$.98 \pm .05}$ & $.69 \pm .23$ & $.75 \pm .17$ & $.87 \pm .12$ & $.62 \pm .24$ \\
        \hline
        \multirow{3}{*}{\rotatebox[origin=c]{90}{\shortstack{Mini\\BooNE}}}
& .1 & $.80 \pm .13$ & \bm{$.94 \pm .08}$ & $.56 \pm .18$ & $.56 \pm .15$ & $.49 \pm .13$ & $.51 \pm .18$ \\
& .3 & $.96 \pm .09$ & \bm{$1.0 \pm .00}$ & $.55 \pm .19$ & $.58 \pm .16$ & $.54 \pm .13$ & $.55 \pm .19$ \\
& .5 & $1.0 \pm .00$ & \bm{$1.0 \pm .00}$ & $.55 \pm .18$ & $.61 \pm .17$ & $.58 \pm .15$ & $.56 \pm .20$ \\
        \hline
        \multirow{3}{*}{\rotatebox[origin=c]{90}{\shortstack{Stat\\log}}}
& .1 & $.77 \pm .17$ & \bm{$.91 \pm .11}$ & $.66 \pm .14$ & $.56 \pm .14$ & $.70 \pm .19$ & $.56 \pm .27$ \\
& .3 & $.96 \pm .09$ & \bm{$.99 \pm .04}$ & $.88 \pm .10$ & $.82 \pm .15$ & \bm{$.95 \pm .07}$ & $.67 \pm .26$ \\
& .5 & $.98 \pm .05$ & \bm{$1.0 \pm .00}$ & $.95 \pm .03$ & $.91 \pm .10$ & \bm{$.99 \pm .03}$ & $.82 \pm .21$ \\
        \hline
        \\
		\multicolumn{8}{c}{(iv) Covariance Change (Conditional)} \\
		& $c$ & \shortstack{Proposed\\(exact)} & Proposed & MT & Id\'e'09 & Hara'15 & SPARDA \\
		\hline
		\multirow{3}{*}{\rotatebox[origin=c]{90}{CASP}}
& .1 & $.64 \pm .21$ & \bm{$.64 \pm .22}$ & \bm{$.52 \pm .15}$ & \bm{$.51 \pm .24}$ & \bm{$.53 \pm .24}$ & \bm{$.63 \pm .15}$ \\
& .3 & $.82 \pm .18$ & \bm{$.82 \pm .20}$ & $.61 \pm .20$ & $.52 \pm .24$ & $.63 \pm .22$ & $.67 \pm .19$ \\
& .5 & $.89 \pm .14$ & \bm{$.92 \pm .15}$ & $.69 \pm .19$ & $.55 \pm .25$ & $.70 \pm .22$ & $.63 \pm .16$ \\
		\hline
		\multirow{3}{*}{\rotatebox[origin=c]{90}{CBM}}
& .1 & $.72 \pm .17$ & \bm{$.76 \pm .18}$ & $.54 \pm .20$ & $.50 \pm .19$ & $.57 \pm .19$ & $.51 \pm .09$ \\
& .3 & $.86 \pm .16$ & \bm{$.92 \pm .11}$ & $.59 \pm .24$ & $.51 \pm .20$ & $.66 \pm .16$ & $.55 \pm .14$ \\
& .5 & $.92 \pm .12$ & \bm{$.95 \pm .09}$ & $.67 \pm .18$ & $.51 \pm .21$ & $.72 \pm .15$ & $.57 \pm .17$ \\
        \hline
        \multirow{3}{*}{\rotatebox[origin=c]{90}{\shortstack{Diag\\nosis}}}
& .1 & $.64 \pm .12$ & \bm{$.74 \pm .16}$ & \bm{$.65 \pm .18}$ & $.57 \pm .17$ & $.55 \pm .19$ & $.42 \pm .14$ \\
& .3 & $.88 \pm .10$ & \bm{$.95 \pm .07}$ & $.64 \pm .25$ & $.64 \pm .19$ & $.68 \pm .17$ & $.47 \pm .15$ \\
& .5 & $.93 \pm .09$ & \bm{$.98 \pm .04}$ & $.64 \pm .24$ & $.67 \pm .20$ & $.77 \pm .14$ & $.50 \pm .17$ \\
        \hline
        \multirow{3}{*}{\rotatebox[origin=c]{90}{\shortstack{Mini\\BooNE}}}
& .1 & $.70 \pm .15$ & \bm{$.76 \pm .17}$ & $.54 \pm .19$ & $.55 \pm .15$ & $.48 \pm .13$ & $.53 \pm .17$ \\
& .3 & $.76 \pm .13$ & \bm{$.90 \pm .11}$ & $.51 \pm .19$ & $.56 \pm .15$ & $.53 \pm .13$ & $.51 \pm .15$ \\
& .5 & $.88 \pm .11$ & \bm{$.95 \pm .09}$ & $.53 \pm .18$ & $.57 \pm .16$ & $.55 \pm .13$ & $.48 \pm .16$ \\
        \hline
        \multirow{3}{*}{\rotatebox[origin=c]{90}{\shortstack{Stat\\log}}}
& .1 & $.57 \pm .14$ & \bm{$.63 \pm .24}$ & \bm{$.48 \pm .22}$ & \bm{$.54 \pm .14}$ & \bm{$.52 \pm .21}$ & \bm{$.58 \pm .23}$ \\
& .3 & $.76 \pm .16$ & \bm{$.83 \pm .20}$ & \bm{$.73 \pm .17}$ & $.67 \pm .18$ & $.68 \pm .17$ & $.62 \pm .22$ \\
& .5 & $.89 \pm .13$ & \bm{$.93 \pm .10}$ & $.87 \pm .09$ & $.74 \pm .16$ & $.76 \pm .16$ & $.64 \pm .23$ \\
        \hline
	\end{tabular}
\end{table*}

\begin{table*}[p]
	\caption{Average AUROC $\pm$ standard deviation on 20 random data realizations from five UCI datasets. Proposed (exact) is a referential result with the exact solution of (\ref{eq:kss}) derived using IBM ILOG CPLEX. The highest AUROC of the five methods is shown in bold. The best results and other results were compared using a t-test (5\%), and results that were not rejected are also highlighted.}
	\label{tab:result_uci3}
	\centering
	\small
	\begin{tabular}{cc|c|ccccc}
		\multicolumn{8}{c}{(v) Covariance Change (No Variance Change)} \\
		& $c$ & \shortstack{Proposed\\(exact)} & Proposed & MT & Id\'e'09 & Hara'15 & SPARDA \\
		\hline
		\multirow{3}{*}{\rotatebox[origin=c]{90}{CASP}}
& .1 & $.60 \pm .17$ & \bm{$.61 \pm .25}$ & \bm{$.68 \pm .15}$ & $.53 \pm .24$ & \bm{$.63 \pm .23}$ & \bm{$.59 \pm .17}$ \\
& .3 & $.87 \pm .16$ & \bm{$.90 \pm .12}$ & $.71 \pm .15$ & $.57 \pm .25$ & \bm{$.85 \pm .14}$ & $.69 \pm .17$ \\
& .5 & $.92 \pm .11$ & \bm{$.95 \pm .07}$ & $.79 \pm .17$ & $.61 \pm .26$ & \bm{$.92 \pm .11}$ & $.72 \pm .14$ \\
		\hline
		\multirow{3}{*}{\rotatebox[origin=c]{90}{CBM}}
& .1 & $.88 \pm .13$ & \bm{$.87 \pm .12}$ & $.57 \pm .22$ & $.50 \pm .19$ & $.69 \pm .14$ & $.51 \pm .15$ \\
& .3 & $.99 \pm .05$ & \bm{$.97 \pm .09}$ & $.68 \pm .23$ & $.52 \pm .20$ & $.82 \pm .12$ & $.58 \pm .14$ \\
& .5 & $.96 \pm .09$ & \bm{$.95 \pm .10}$ & $.80 \pm .17$ & $.53 \pm .22$ & $.83 \pm .12$ & $.73 \pm .12$ \\
        \hline
        \multirow{3}{*}{\rotatebox[origin=c]{90}{\shortstack{Diag\\nosis}}}
& .1 & $.58 \pm .14$ & \bm{$.71 \pm .15}$ & $.58 \pm .23$ & \bm{$.63 \pm .15}$ & $.58 \pm .19$ & $.43 \pm .13$ \\
& .3 & $.84 \pm .14$ & \bm{$.92 \pm .10}$ & $.59 \pm .26$ & $.69 \pm .17$ & $.79 \pm .13$ & $.47 \pm .16$ \\
& .5 & $.89 \pm .12$ & \bm{$.94 \pm .09}$ & $.68 \pm .24$ & $.75 \pm .17$ & $.87 \pm .12$ & $.61 \pm .18$ \\
        \hline
        \multirow{3}{*}{\rotatebox[origin=c]{90}{\shortstack{Mini\\BooNE}}}
& .1 & $.79 \pm .13$ & \bm{$.93 \pm .09}$ & $.55 \pm .18$ & $.56 \pm .15$ & $.49 \pm .13$ & $.52 \pm .17$ \\
& .3 & $.96 \pm .09$ & \bm{$1.0 \pm .00}$ & $.55 \pm .19$ & $.58 \pm .16$ & $.54 \pm .13$ & $.53 \pm .20$ \\
& .5 & $1.0 \pm .00$ & \bm{$1.0 \pm .00}$ & $.57 \pm .18$ & $.61 \pm .17$ & $.58 \pm .15$ & $.56 \pm .19$ \\
        \hline
        \multirow{3}{*}{\rotatebox[origin=c]{90}{\shortstack{Stat\\log}}}
& .1 & $.77 \pm .18$ & \bm{$.90 \pm .11}$ & $.70 \pm .18$ & $.56 \pm .14$ & $.70 \pm .19$ & $.58 \pm .22$ \\
& .3 & $.95 \pm .10$ & \bm{$.98 \pm .07}$ & $.89 \pm .10$ & $.82 \pm .15$ & \bm{$.95 \pm .07}$ & $.66 \pm .26$ \\
& .5 & $.98 \pm .05$ & \bm{$1.0 \pm .00}$ & $.95 \pm .04$ & $.91 \pm .10$ & \bm{$.99 \pm .03}$ & $.82 \pm .19$ \\
	\end{tabular}
	\caption{Average runtime $\pm$ standard deviation of the proposed method and SPARDA with ten-thread parallelization on the first two changes. The smaller runtime is highlighted.}
	\label{tab:result_uci_time}
	\centering
	\small
	\subfigure{
	\begin{tabular}{cc|cc}
		\multicolumn{4}{c}{(i) Mean Shift} \\
		& $c$ & Proposed & SPARDA \\
		\hline
		\multirow{3}{*}{\rotatebox[origin=c]{90}{CASP}}
& .1 & \bm{$0.13 \pm 0.00}$ & $13.4 \pm 7.32$ \\
& .3 & \bm{$0.13 \pm 0.00}$ & $5.13 \pm 2.39$ \\
& .5 & \bm{$0.14 \pm 0.03}$ & $2.31 \pm 0.78$ \\
		\hline
		\multirow{3}{*}{\rotatebox[origin=c]{90}{CBM}}
& .1 & \bm{$0.18 \pm 0.05}$ & $25.8 \pm 11.7$ \\
& .3 & \bm{$0.18 \pm 0.05}$ & $5.24 \pm 1.87$ \\
& .5 & \bm{$0.19 \pm 0.05}$ & $2.03 \pm 0.51$ \\
        \hline
        \multirow{3}{*}{\rotatebox[origin=c]{90}{\shortstack{Diag\\nosis}}}
& .1 & \bm{$0.77 \pm 0.05}$ & $29.5 \pm 48.2$ \\
& .3 & \bm{$0.77 \pm 0.05}$ & $28.3 \pm 49.6$ \\
& .5 & \bm{$0.76 \pm 0.04}$ & $26.8 \pm 49.5$ \\
        \hline
        \multirow{3}{*}{\rotatebox[origin=c]{90}{\shortstack{Mini\\BooNE}}}
& .1 & \bm{$0.83 \pm 0.06}$ & $127 \pm 57.1$ \\
& .3 & \bm{$0.81 \pm 0.04}$ & $126 \pm 55.7$ \\
& .5 & \bm{$0.82 \pm 0.05}$ & $123 \pm 55.6$ \\
        \hline
        \multirow{3}{*}{\rotatebox[origin=c]{90}{\shortstack{Stat\\log}}}
& .1 & \bm{$0.53 \pm 0.02}$ & $12.9 \pm 6.03$ \\
& .3 & \bm{$0.52 \pm 0.03}$ & $4.62 \pm 1.34$ \\
& .5 & \bm{$0.52 \pm 0.03}$ & $2.02 \pm 0.45$ \\
	\end{tabular}
	}
		\subfigure{
	\begin{tabular}{cc|cc}
		\multicolumn{4}{c}{(ii) Variance Change} \\
		& $c$ & Proposed & SPARDA \\
		\hline
		\multirow{3}{*}{\rotatebox[origin=c]{90}{CASP}}
& .1 & \bm{$0.13 \pm 0.02}$ & $13.9 \pm 7.08$ \\
& .3 & \bm{$0.13 \pm 0.00}$ & $14.4 \pm 8.23$ \\
& .5 & \bm{$0.13 \pm 0.02}$ & $13.5 \pm 6.21$ \\
		\hline
		\multirow{3}{*}{\rotatebox[origin=c]{90}{CBM}}
& .1 & \bm{$0.17 \pm 0.05}$ & $43.1 \pm 12.9$ \\
& .3 & \bm{$0.17 \pm 0.05}$ & $36.1 \pm 11.0$ \\
& .5 & \bm{$0.20 \pm 0.04}$ & $21.4 \pm 7.19$ \\
        \hline
        \multirow{3}{*}{\rotatebox[origin=c]{90}{\shortstack{Diag\\nosis}}}
& .1 & \bm{$0.77 \pm 0.05}$ & $27.8 \pm 45.8$ \\
& .3 & \bm{$0.75 \pm 0.03}$ & $29.6 \pm 46.5$ \\
& .5 & \bm{$0.75 \pm 0.04}$ & $31.1 \pm 50.2$ \\
        \hline
& .1 & \bm{$0.82 \pm 0.04}$ & $128 \pm 70.6$ \\
& .3 & \bm{$0.81 \pm 0.04}$ & $131 \pm 79.0$ \\
& .5 & \bm{$0.82 \pm 0.05}$ & $129 \pm 78.5$ \\
        \hline
        \multirow{3}{*}{\rotatebox[origin=c]{90}{\shortstack{Stat\\log}}}
& .1 & \bm{$0.52 \pm 0.03}$ & $14.9 \pm 6.39$ \\
& .3 & \bm{$0.52 \pm 0.03}$ & $14.0 \pm 5.77$ \\
& .5 & \bm{$0.52 \pm 0.04}$ & $16.2 \pm 8.04$ \\
	\end{tabular}
	}
\end{table*}

Tables~\ref{tab:result_uci1}--\ref{tab:result_uci3} show the results on three difference levels $c = 0.1, 0.3$, and $0.5$ over 20 random data realizations.
From the tables, we find two important results that show the effectiveness of the proposed method.
The first finding is that the AUROC of the proposed method attained the best average score among the five methods for almost all cases.
Moreover, we observe that there is more than 0.2 improvement in the average AUROCs of the proposed method compared with those of the Gaussian-based methods for some cases.
As discussed in Section~\ref{sec:example}, this is because the proposed method can detect a complex distribution difference more effectively than the Gaussian-based methods.
Note that the proposed method also outperformed SPARDA.
We conjecture that this was because SPARDA tended to be trapped by local optima when solving the nonconvex optimization.

The second finding exists in the left two columns.
The results show that the proposed method with the greedy scoring method attained comparable results with the exact solution of the sparsest $k$-subgraph problem (\ref{eq:kss}).
In other words, the greedy scoring method (Algorithm~\ref{alg:greedy_score}) provided good approximate solutions and can be a practical alternative for the exact method, which may require exponential time.
Note that the greedy scoring method sometimes outperformed the exact method.
This is because the exact method scores each feature with 0 or 1.
In the exact method, if one feature is misspecified (i.e., scored as 0 instead of 1), that feature is ranked equally to the other features with no distribution differences.
This induces a substantial decrease in the AUROC because only the order of the scores is important when it is computed.
By contrast, the greedy scoring method is less sensitive to such a misspecification.
Some features may be scored lower than the ideal because of a misspecification, but the score of such features can still remain a bit high and thus tend to remain at a higher order than the other features with no distribution differences.
Hence, the decrease in the AUROC is limited.

\tablename~\ref{tab:result_uci_time} shows the computational efficiency of the proposed method.
In the UCI dataset experiments, the runtime of the proposed method was from 3 to more than 100 times faster than the entire runtime of SPARDA.

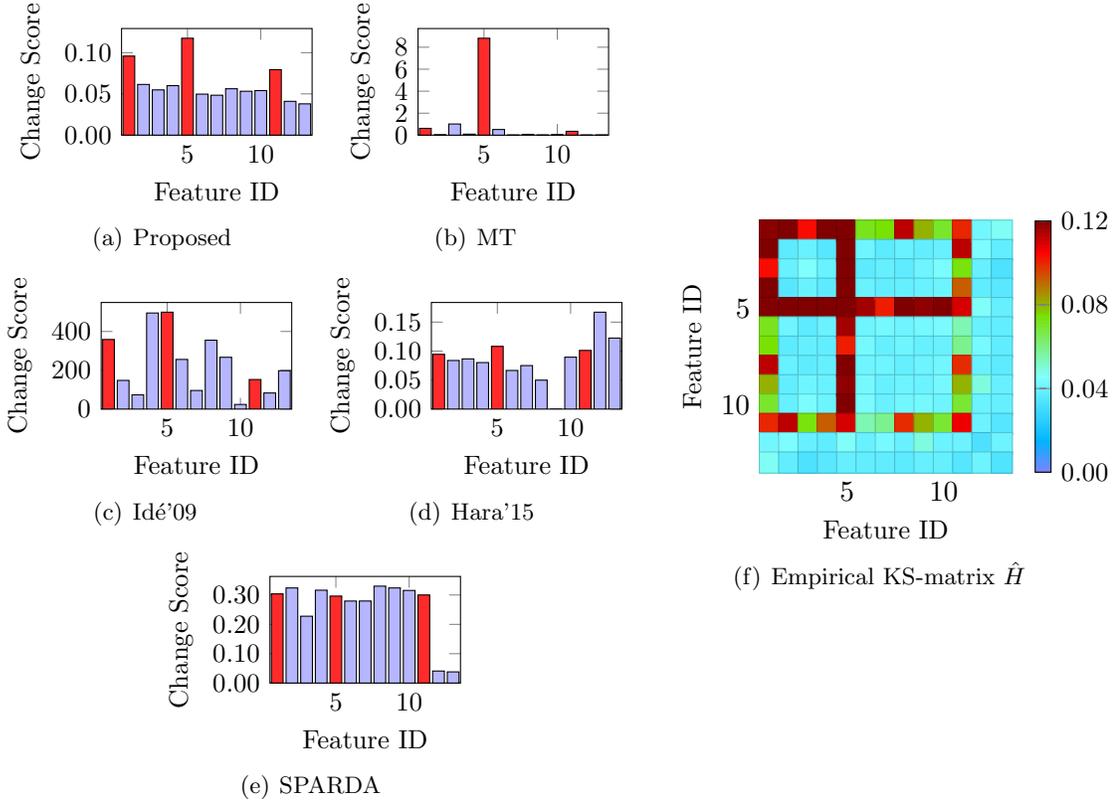
\begin{figure}[t]
	\centering
	\begin{minipage}{0.60\linewidth}
	\subfigure[Proposed]{
	\begin{tikzpicture}
	\begin{axis}[
		xmin=0.5,xmax=13.5,
		ymin=0.0,
		width=0.45\linewidth,
		height=3.0cm,
		xlabel={\small Feature ID},
		ylabel={\small Change Score},
		xticklabel style={font=\small,},
		yticklabel style={
			font=\small,
			/pgf/number format/.cd,
			fixed,
			fixed zerofill,
			precision=2,},
		]
	\addplot[ybar, fill={rgb:white,10;blue,4}, bar width=0.8] table[
		x=dim,y=NAL2,col sep=comma,
		skip coords between index={0}{1},skip coords between index={4}{5},skip coords between index={10}{11}
		]{cbm_q30_k03_r01_scoreC.csv};
	\addplot[ybar, fill={rgb:white,1;red,5}, bar width=0.8] table[
		x=dim,y=NAL2,col sep=comma,
		skip coords between index={1}{4},skip coords between index={5}{10},skip coords between index={11}{13}
		]{cbm_q30_k03_r01_scoreC.csv};
	\end{axis}
	\end{tikzpicture}
	\label{fig:pal}
	}
	\subfigure[MT]{
	\begin{tikzpicture}
	\begin{axis}[
		xmin=0.5,xmax=13.5,
		ymin=0.0,
		width=0.45\linewidth,
		height=3.0cm,
		xlabel={\small Feature ID},
		ylabel={\small Change Score},
		xticklabel style={font=\small,},
		yticklabel style={
			font=\small,
			/pgf/number format/.cd,
			fixed,
			fixed zerofill,
			precision=0,},
		]
	\addplot[ybar, fill={rgb:white,10;blue,4}, bar width=0.8] table[
		x=dim,y=MT,col sep=comma,
		skip coords between index={0}{1},skip coords between index={4}{5},skip coords between index={10}{11}
		]{cbm_q30_k03_r01_scoreC.csv};
	\addplot[ybar, fill={rgb:white,1;red,5}, bar width=0.8] table[
		x=dim,y=MT,col sep=comma,
		skip coords between index={1}{4},skip coords between index={5}{10},skip coords between index={11}{13}
		]{cbm_q30_k03_r01_scoreC.csv};
	\end{axis}
	\end{tikzpicture}
	\label{fig:mt}
	}
	\subfigure[Id\'e'09]{
	\begin{tikzpicture}
	\begin{axis}[
		xmin=0.5,xmax=13.5,
		ymin=0.0,
		width=0.45\linewidth,
		height=3.0cm,
		xlabel={\small Feature ID},
		ylabel={\small Change Score},
		xticklabel style={font=\small,},
		yticklabel style={
			font=\small,
			/pgf/number format/.cd,
			fixed,
			fixed zerofill,
			precision=0,},
		]
	\addplot[ybar, fill={rgb:white,10;blue,4}, bar width=0.8] table[
		x=dim,y=Ide,col sep=comma,
		skip coords between index={0}{1},skip coords between index={4}{5},skip coords between index={10}{11}
		]{cbm_q30_k03_r01_scoreC.csv};
	\addplot[ybar, fill={rgb:white,1;red,5}, bar width=0.8] table[
		x=dim,y=Ide,col sep=comma,
		skip coords between index={1}{4},skip coords between index={5}{10},skip coords between index={11}{13}
		]{cbm_q30_k03_r01_scoreC.csv};
	\end{axis}
	\end{tikzpicture}
	\label{fig:ide}
	}
	\centering
	\subfigure[Hara'15]{
	\begin{tikzpicture}
	\begin{axis}[
		xmin=0.5,xmax=13.5,
		ymin=0.0,
		width=0.45\linewidth,
		height=3.0cm,
		xlabel={\small Feature ID},
		ylabel={\small Change Score},
		xticklabel style={font=\small,},
		yticklabel style={
			font=\small,
			/pgf/number format/.cd,
			fixed,
			fixed zerofill,
			precision=2,},
		]
	\addplot[ybar, fill={rgb:white,10;blue,4}, bar width=0.8] table[
		x=dim,y=Hara,col sep=comma,
		skip coords between index={0}{1},skip coords between index={4}{5},skip coords between index={10}{11}
		]{cbm_q30_k03_r01_scoreC.csv};
	\addplot[ybar, fill={rgb:white,1;red,5}, bar width=0.8] table[
		x=dim,y=Hara,col sep=comma,
		skip coords between index={1}{4},skip coords between index={5}{10},skip coords between index={11}{13}
		]{cbm_q30_k03_r01_scoreC.csv};
	\end{axis}
	\end{tikzpicture}
	\label{fig:hara}
	}
	\subfigure[SPARDA]{
	\begin{tikzpicture}
	\begin{axis}[
		xmin=0.5,xmax=13.5,
		ymin=0.0,
		width=0.45\linewidth,
		height=3.0cm,
		xlabel={\small Feature ID},
		ylabel={\small Change Score},
		xticklabel style={font=\small,},
		yticklabel style={
			font=\small,
			/pgf/number format/.cd,
			fixed,
			fixed zerofill,
			precision=2,},
		]
	\addplot[ybar, fill={rgb:white,10;blue,4}, bar width=0.8] table[
		x=dim,y=SPARDA,col sep=comma,
		skip coords between index={0}{1},skip coords between index={4}{5},skip coords between index={10}{11}
		]{cbm_q30_k03_r01_scoreC.csv};
	\addplot[ybar, fill={rgb:white,1;red,5}, bar width=0.8] table[
		x=dim,y=SPARDA,col sep=comma,
		skip coords between index={1}{4},skip coords between index={5}{10},skip coords between index={11}{13}
		]{cbm_q30_k03_r01_scoreC.csv};
	\end{axis}
	\end{tikzpicture}
	\label{fig:sparda}
	}
	\end{minipage}
	\begin{minipage}{0.34\textwidth}
	\centering
	\subfigure[Empirical KS-matrix $\hat{H}$]{
	\begin{tikzpicture}
	\begin{axis}[
		width=0.95\textwidth,
		height=0.95\textwidth,
		tick align=inside,
		unbounded coords=jump,
		xmin=0.5,
		xmax=13.5,
		ymin=0.5,
		ymax=13.5,
		xlabel={\small Feature ID},
		ylabel={\small Feature ID},
		y dir=reverse,
		point meta min=0,
		point meta max=0.12,
		colorbar,
		colormap={jet}{rgb255(0cm)=(128,128,255) rgb255(1cm)=(0,180,255) rgb255(3cm)=(128,255,255) rgb255(5cm)=(120,225,0) rgb255(7cm)=(255,0,0) rgb255(8cm)=(128,0,0)},
		colorbar style={
			width=6pt,
			ytick={0,0.04,0.08,0.12},
			yticklabel style={
				font=\small,
				text width=0.5em,
				align=right,
				/pgf/number format/.cd,
				fixed,
				fixed zerofill
				},
			},
		]
		\addplot[mark=square*,only marks, scatter, scatter src=explicit, mark size=4]
		file {cbm_q30_k03_r01_scoreC_L.csv};
	\end{axis}
	\end{tikzpicture}
	\label{fig:H}
	}
	\end{minipage}
	\caption{Results on  the CBM dataset with Covariance Change ($c=0.3$): (a)--(e) Change score: red bars on the 1st, 5th, and 11th features denote that they are features with distribution differences, while blue bars on the other features denote that they have no distribution differences. (f) Empirical KS-matrix $\hat{H}$. }
	\label{fig:scores}
\end{figure}

To demonstrate the success of the proposed method in detail, we show a result from the CBM dataset with Covariance Change ($c=0.3$) in \figurename~\ref{fig:scores}.
In this example, we set the features with distribution differences as $S^* = \{1, 5, 11\}$.
In \figurename~\ref{fig:pal}, we observe that the score of the proposed method marked the top-three values on the set $S^*$, which is an ideal result.
This is not the case with the other four baseline methods.
The two Gaussian-based methods MT and Id\'e'09 marked the largest score on the fifth feature, but they failed to detect the other two features, whereas Hara'15 marked the largest score on the twelfth feature, which does not have distribution differences.
SPARDA marked nearly equal scores for the first ten features and, hence, failed to detect features with distribution differences.
The empirical KS-matrix $\hat{H}$ in \figurename~\ref{fig:H} shows why the proposed method could detect differences successfully.
Matrix $\hat{H}$ had large values on the rows and columns that correspond to set $S^*$.
This means that Conditions~(S1) and (S2) in Theorem~\ref{th:consistency_suff} are met; thus, the set $S^*$ was detected properly.

\subsection{Application to Anomaly Detection in Quantum Systems}
\label{sec:quantum}

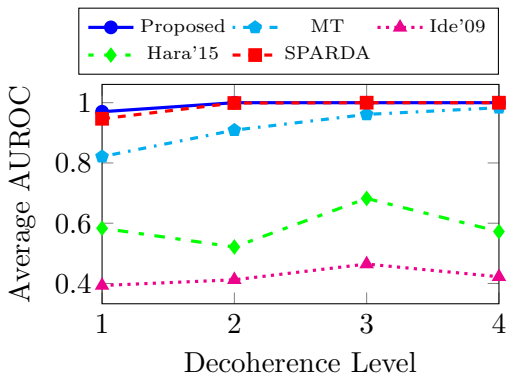
\begin{figure}[t]
	\begin{minipage}{0.48\textwidth}
	\centering
	\tblcaption{Decoherence level as well as the mean and variance of the $(1, 4)$-th entry. Level 0 is normal data.}
	\label{tab:quantum}
	\begin{tabular}{c|cc}
		Level & Mean & Variance \\
		\hline
0 & 0.42 & 0.000651\\
\hline
1 & 0.40 & 0.001055\\
2 & 0.38 & 0.000876\\
3 & 0.36 & 0.000817\\
4 & 0.34 & 0.000768\\
	\end{tabular}
	\end{minipage}
	\hfill
	\begin{minipage}{0.5\textwidth}
	\begin{tikzpicture}
	\begin{axis}[
		scale=1.0,
		xlabel={Decoherence Level},
		xmin=1,
		xmax=4,
		ylabel style={align=center},
		ylabel={Average AUROC},
		width=0.90\textwidth,
		height=4.5cm,
		legend style={at={(1.00,1.35)}, font=\fontsize{7}{5}\selectfont},
		legend columns=3,
	]
	\addplot [color=blue, line width=1.2pt, mark=*, mark options={solid, scale=1.0, fill=blue}, solid]
	table [x=level,y=NAL2,col sep=comma]{pra_real.csv};
	\addplot [color=cyan, line width=1.2pt, mark=pentagon*, mark options={solid, scale=1.0, fill=cyan}, dash pattern=on 1pt off 3pt on 3pt off 3pt]
	table [x=level,y=MT,col sep=comma]{pra_real.csv};
	\addplot [color=magenta, line width=1.2pt, mark=triangle*, mark options={solid, scale=1.0, fill=magenta}, dotted]
	table [x=level,y=Ide,col sep=comma]{pra_real.csv};
	\addplot [color=green, line width=1.2pt, mark=diamond*, mark options={solid, scale=1.0, fill=green}, dash pattern=on 2pt off 4pt on 4pt off 4pt]
	table [x=level,y=Hara,col sep=comma]{pra_real.csv};
	\addplot [color=red, line width=1.2pt, mark=square*, mark options={solid, scale=1.0, fill=red}, dashed]
	table [x=level,y=SPARDA,col sep=comma]{pra_real.csv};
	\legend{Proposed, MT, Ide'09, Hara'15, SPARDA}
	\end{axis}
	\end{tikzpicture}
	\caption{AUROC on decoherence data}
	\label{fig:quantum}
	\end{minipage}
\end{figure}

We applied the proposed method to anomaly detection in quantum systems~\citep{hara2014anomaly,hara2016quantum}.
In quantum informatics, we sometimes face unknown errors in the given quantum state.
For such cases, it is critically important to find the error sources for several applications, such as quantum computation, quantum cryptography, and quantum metrology.

In this experiment, we used data derived from a real physical experiment.
In the physical experiment, 300 normal density matrices were derived, each of which is a $4 \times 4 $ Hermitian matrix.
50 erroneous matrices were also derived with a decoherence in their $(1, 4)$-th entry.
Appendix~\ref{sec:qsupp} lists the details of the experimental settings.
Experimentally obtained density matrices have changes in both the mean and variance on the $(1, 4)$-th entry (\tablename~\ref{tab:quantum}). 
Here, the task is to find the erroneous $(1, 4)$-th entry using different-feature selection. 

Before the experiment, we applied two preprocessing steps.
First, because the error appears only on the absolute value of the matrix entry, we computed the absolute value of each entry.
Second, because the matrix is symmetric, we extracted only the upper-triangular entries and transformed the matrix into a ten-dimensional vector.

In the experiment, we randomly sampled $N = M = 25$ vectors from both normal and erroneous data, and then applied different-feature selection methods.
We repeated this procedure 100 times.

\figurename~\ref{fig:quantum} shows the average AUROC over 100 random data realizations for each method.
It indicates that the Gaussian-based methods performed poorly compared to the proposed method and SPARDA.
The proposed method and SPARDA attained AUROC=1 except when the decoherence level was one. 
To examine the performance difference of these two methods in detail, we applied a t-test to the AUROCs of these two methods under a decoherence level of one. 
The result of the t-test rejected the null-hypothesis (i.e., that their average performances are equal) at a 5\% p-value. 
This means that the proposed method could find the different-features most effectively.

\section{Conclusion}
\label{sec:concl}

We proposed a simple nonparametric method for different-feature selection that satisfies two requirements, namely, less restrictive assumptions on the distributions and computational efficiency.
In the proposed method, we first computed the empirical KS-matrix and then solved the sparsest $k$-subgraph problem derived from the matrix using a greedy scoring method.
We showed that the proposed method runs in only $O(D^2 L N \log N)$ time.
Moreover, it does not require extra computation for model selection.
We also proved that the proposed method provides a consistent solution under mild conditions.
In particular, it requires less restrictive assumptions on the data distributions for consistent estimation than the current Gaussian-based methods.

The experimental results revealed that the proposed method significantly outperformed the Gaussian-based methods.
The proposed method detected the complex distribution difference effectively and attained a high AUROC even for cases in which the Gaussian-based methods worked poorly.
We also compared the proposed method to the state-of-the-art SPARDA method.
The experimental results showed that the proposed method attained a higher AUROC than SPARDA on several datasets while requiring less computation time.

Despite the computational efficiency of the proposed method, there still remains a scalability issue, that is, the time complexity is proportional to $D^2$, which can be prohibitive in a high dimensional setting.
Improving the computational scalability is one direction of our future work.

\section*{Acknowledgements}
This work was supported by JST ERATO Grant Number JPMJER1201, Japan.
This work was also supported in part by JST CREST Grant Number JPMJCR1304, Japan.
This work was also supported in part by JSPS KAKENHI (17J03208), Japan.
This work was also supported in part by the National Science Foundation (NSF grant IIS-9988642), the Multidisciplinary Research Program of the Department of Defense (MURI N00014-00-1-0637), JST-CREST project (JPMJCR1674), and Grant-in-Aid from JSPS (26220712).

\appendix
\section{Baseline Methods}
\label{sec:baseline}

We present the details of the baseline methods in Section~\ref{sec:simu}.

\paragraph{Notation:}
$\bm{\mu}_{\mathcal{P}}, \bm{\mu}_{\mathcal{Q}} \in \R^d$ and $\Sigma_{\mathcal{P}}, \Sigma_{\mathcal{Q}} \in \R^{D \times D}$ denote the empirical averages and covariances of datasets $\mathcal{P}$ and $\mathcal{Q}$, respectively.
Moreover, $\Lambda_{\mathcal{P}}$ and $\Lambda_{\mathcal{Q}} \in \R^{D \times D}$ denote the estimated precision matrices of datasets $\mathcal{P}$ and $\mathcal{Q}$ using the Tikhonov-regularization method, respectively.
That is, we define $\Lambda_{\mathcal{P}} := (\Sigma_{\mathcal{P}} + \kappa I_D)^{-1}$ and $\Lambda_{\mathcal{Q}} := (\Sigma_{\mathcal{Q}} + \kappa I_D)^{-1}$ with a regularization parameter $\kappa$.
The value of $\kappa$ is chosen from 11 different parameter candidates between $10^{-4}$ and $10^{1}$ using three-fold cross validation.
For a square matrix $U \in \R^{D \times D}$ and a set $S \subseteq [D]$, we denote the submatrix by $U_S := \{U_{ij} \mid i, j \in S\}$.

\paragraph{[MT~\citep{taguchi2000new}]}
We adopted a simplified version of MT for ease of computation.
We used a combinatorial optimization instead of the F-test in the original MT.
The estimated feature set $\hat{S}_{\rm MT}$ is given by solving the next problem:
\begin{align}
	\begin{split}
	\hat{S}_{\rm MT}^\comp &= \argminl_{S^\comp \subseteq [D]} \left| D - \alpha - \tr{\Gamma_{S^\comp} C_{S^\comp}^{-1}} \right| , \\
	& \text{subject to} \; |S^\comp| = k ,
	\end{split}
	\label{eq:mt}
\end{align}
where $\Gamma := \frac{1}{M} \sum_{m=1}^M (\bm{z}^{(m)} - \bm{\mu}_{\mathcal{P}}) (\bm{z}^{(m)} - \bm{\mu}_{\mathcal{P}})^\top$ and $C := \Lambda_{\mathcal{P}}^{-1}$.
Because the number $k$ is unknown, we used the greedy scoring method (Algorithm~\ref{alg:greedy_score}) to solve problem (\ref{eq:mt}), where we defined $f(S) := \left| |S^\comp| - \tr{\Gamma_{S^\comp} C_{S^\comp}^{-1}} \right|$.

\paragraph{[Id\'e'09~\citep{ide2009proximity}]}
In Id\'e'09, the score of the $d$-th feature $\hat{s}_d$ is given by
\begin{align}
	\hat{s}_d :=& \, \max \{\hat{s}_d^{\mathcal{P} \mathcal{Q}}, \hat{s}_d^{\mathcal{Q} \mathcal{P}}\} , \\
	\hat{s}_d^{\mathcal{P} \mathcal{Q}} :=& \, \bm{w}_{\mathcal{P}}^\top (\bm{\ell}_{\mathcal{Q}} - \bm{\ell}_{\mathcal{P}}) + \frac{1}{2} \left\{ \frac{\bm{\ell}_{\mathcal{Q}}^\top W_{\mathcal{P}} \bm{\ell}_{\mathcal{Q}}}{\lambda_{\mathcal{Q}}} - \frac{\bm{\ell}_{\mathcal{P}}^\top W_{\mathcal{P}} \bm{\ell}_{\mathcal{P}}}{\lambda_{\mathcal{P}}} \right\} + \frac{1}{2} \left\{ \log \frac{\lambda_{\mathcal{P}}}{\lambda_{\mathcal{Q}}} + \sigma_{\mathcal{P}} (\lambda_{\mathcal{P}} - \lambda_{\mathcal{Q}}) \right\} ,
\end{align}
where the matrices are partitioned as
\begin{align*}
	\Lambda_{\mathcal{P}} = \matrix{L_{\mathcal{P}} & \bm{\ell}_{\mathcal{P}} \\ \bm{\ell}_{\mathcal{P}}^\top & \lambda_{\mathcal{P}}}, \; \Lambda_{\mathcal{P}}^{-1} = \matrix{W_{\mathcal{P}} & \bm{w}_{\mathcal{P}} \\ \bm{w}_{\mathcal{P}}^\top & \sigma_{\mathcal{P}}} .
\end{align*}
Here, we assume that the rows and columns of $\Lambda_{\mathcal{P}}$ and $\Lambda_{\mathcal{P}}^{-1}$ are permuted so that their original $d$-th rows and columns are located at the last rows and columns of the matrix.
Matrices $\Lambda_{\mathcal{Q}}$ and $\Lambda_{\mathcal{Q}}^{-1}$ are partitioned in the same manner.

\paragraph{[Hara'15~\citep{hara2015consistent}]}
Similarly to the proposed method, Hara'15 uses the sparsest $k$-subgraph problem~(\ref{eq:kss}).
Matrix $\hat{H}$ is given by $\hat{H}_{ij} := |\Sigma_{\mathcal{P}, ij} - \Sigma_{\mathcal{Q}, ij}|$.
We used the greedy scoring method (Algorithm~\ref{alg:greedy_score}) to solve the problem.

\paragraph{[SPARDA~\citep{mueller2015principal}]}
The solution of SPARDA $\hat{\bm{\beta}}$ can be derived by solving the max-min problem:
\begin{align}
	\max_{\bm{\beta} \in \mathcal{B}} \min_{U \in \mathcal{M}} \sum_{n=1}^N \sum_{m=1}^M (\bm{\beta}^\top \bm{y}^{(n)} - \bm{\beta}^\top \bm{z}^{(m)})^2 U_{nm} - \lambda \|  \bm{\beta}\|_1 ,
\end{align}
where $\mathcal{B} := \{\bm{\beta} \in \R^D \mid \|\bm{\beta}\| \leq 1, \beta_1 \geq 0\}$ and $\mathcal{M} = \{U \in \R_+^{N \times M} \mid \forall m, \sum_{n=1}^N U_{nm} = 1/M \; {\rm and} \; \forall n, \sum_{m=1}^M U_{nm} = 1/N\}$.
The minimization term corresponds to computing the Wasserstein distance between the distributions.
We implemented SPARDA using C++ based on the MATLAB code \texttt{fastSPARDA.m} available on the author's website~\footnote{\url{http://www.mit.edu/~jonasm/}}.
Because the relax and tighten procedure proposed by \cite{mueller2015principal} was too slow, we used the projected gradient ascent, which runs in $O(DN + N \log N)$ per iteration.
In our preliminary experiment, we observed that the projected gradient ascent ran more than ten times faster than the relax and tighten procedure.
Because the projected gradient ascent tends to be trapped by local optima, we used five random restarts.
We set the parameter candidate for $\lambda$ to $\{0, 10^{-4}, 10^{-3}, 10^{-2}, 10^{-1}\}$ and selected the optimal one using five-fold cross validation.
After we derived solution $\hat{\bm{\beta}}$, we set the score of each feature as $\hat{s}_d = |\hat{\beta}_d|$.

\section{Quantum Data: Experimental Setup}
\label{sec:qsupp}

In order to confirm the performance of the proposed method, we experimentally obtained various density matrices of qubits.
In the experiment, we used a two-photon polarization entangled state for the ``normal state''~\citep{hara2014anomaly}.
For the erroneous states, we prepared several quantum states where the amplitude of the off-diagonal elements of the density matrices slightly vary from the normal state.
Note that the elements of the normal and erroneous density matrices have intrinsic fluctuations because of the limited number of samples (photon pairs) used for reconstruction by Quantum State Tomography (QST)~\citep{Jam2001}.

We used a pair of $\beta$ Barium Borate (BBO) crystals pumped by a continuous wave (CW) diode laser at 405 nm to generate the polarization entangled state $|\psi \rangle = (|H;H \rangle_{a,b} + |V;V \rangle_{a,b})/\sqrt{2}$, where $H$ and $V$ represent horizontally and vertically polarized photons, respectively, and $a$ and $b$ denote spatial modes~\citep{hara2014anomaly,hara2016quantum}.
The measurement outcome of different 16 measurement bases, to each of which approximately 1,000 photon pairs contributed, is converted into density matrix using the conventional QST method~\citep{Jam2001}.
For the density matrices of the erroneous states, we experimentally obtained the measurement outcomes using the three input states $|\psi \rangle, |H;H\rangle_{a,b}$, and $|V;V\rangle_{a,b}$ separately and added them together so that the amplitude of the off-diagonal terms of the density matrices are changed from that of the pure entangled state~\citep{hara2014anomaly,hara2016quantum}. 
For the analysis, 300 normal density matrices and 50 erroneous matrices were derived.
We note that the detection of the change in the (1, 4)-th entry is equivalent to the detection of the change in the quantity of entanglement under the assumption that local polarization does not flip between $|H \rangle$ and $|V \rangle$.

\section{Proofs of the Theorems}
\label{sec:proof}

\subsection{Preliminaries}
We first give three lemmas that we use in the proofs of the theorems.

\begin{lemma}
	\label{lem:lem1}
	The following inequality holds:
	\begin{align}
		\left| {\rm KS}(p_i, q_i) - {\rm KS}(\hat{p}_i, \hat{q}_i) \right| \le \|P_i - \hat{P}_i\|_\infty + \|Q_i - \hat{Q}_i\|_\infty .
	\end{align}
\end{lemma}
\paragraph{Proof}
	Recall the definition of the KS statistic:
	\begin{align*}
		{\rm KS}(p_i, q_i) = \| P_i - Q_i \|_\infty , \qquad {\rm KS}(\hat{p}_i, \hat{q}_i) = \| \hat{P}_i - \hat{Q}_i \|_\infty .
	\end{align*}
	Hence, we have
	\begin{align*}
		{\rm KS}(p_i, q_i) - {\rm KS}(\hat{p}_i, \hat{q}_i) & = \| P_i - Q_i \|_\infty - \| \hat{P}_i - \hat{Q}_i \|_\infty \\
		& \le \| P_i - \hat{P}_i \|_\infty + \| \hat{P}_i - Q_i \|_\infty - \| \hat{P}_i - Q_i \|_\infty + \| Q_i - \hat{Q}_i \|_\infty \\
		& = \|P_i - \hat{P}_i \|_\infty + \|Q_i - \hat{Q}_i \|_\infty .
	\end{align*}
	The opposite direction can be proved in a similar manner:
	\begin{align*}
		 {\rm KS}(\hat{p}_i, \hat{q}_i) - {\rm KS}(p_i, q_i) \le \|P_i - \hat{P}_i \|_\infty + \|Q_i - \hat{Q}_i \|_\infty .
	\end{align*}
\hfill $\Box$

\begin{lemma}
	\label{lem:lem2}
	The following inequality holds:
	\begin{align}
		\left| g(p_{ij}, q_{ij}) - g(\hat{p}_{ij}, \hat{q}_{ij}) \right| \le \sup_{\theta \in [0, \pi]} \left( \|P_{ij, \theta} - \hat{P}_{ij, \theta}\|_\infty + \|Q_{ij, \theta} - \hat{Q}_{ij, \theta}\|_\infty \right) ,
	\end{align}
	where $P_{ij, \theta}$, $\hat{P}_{ij, \theta}$ and $Q_{ij, \theta}$, $\hat{Q}_{ij, \theta}$  are the true and the empirical distribution functions of variable  $r_{ij, \theta} = x_i \cos \theta + x_j \sin \theta$ under distributions $p$ and $q$, respectively.
\end{lemma}
\paragraph{Proof}
	From Lemma~\ref{lem:lem1}, for any $\theta \in [0, \pi], $we have
	\begin{align*}
		 \left| {\rm KS}(p_{ij, \theta}, q_{ij, \theta}) - {\rm KS}(\hat{p}_{ij, \theta}, \hat{q}_{ij, \theta}) \right| \le \|P_{ij, \theta} - \hat{P}_{ij, \theta}\|_\infty + \|Q_{ij, \theta} - \hat{Q}_{ij, \theta}\|_\infty .
	\end{align*}
	We then have
	\begin{align*}
		\left| g(p_{ij}, q_{ij}) - g(\hat{p}_{ij}, \hat{q}_{ij}) \right| & \le \mathbb{E}_{\theta \sim \mathcal{U}(0, \pi)} \left[ \left| {\rm KS}(p_{ij, \theta}, q_{ij, \theta}) - {\rm KS}(\hat{p}_{ij, \theta}, \hat{q}_{ij, \theta}) \right| \right] \\
		& \le \mathbb{E}_{\theta \sim \mathcal{U}(0, \pi)} \left[ \|P_{ij, \theta} - \hat{P}_{ij, \theta}\|_\infty + \|Q_{ij, \theta} - \hat{Q}_{ij, \theta}\|_\infty \right] \\
		& \le \sup_{\theta \in [0, \pi]} \left( \|P_{ij, \theta} - \hat{P}_{ij, \theta}\|_\infty + \|Q_{ij, \theta} - \hat{Q}_{ij, \theta}\|_\infty \right) .
	\end{align*}
\hfill $\Box$

\begin{lemma}
	\label{lem:lem3}
	There exists $A_{ij}', B_{ij}' > 0$ such that, for any $\tau > 0$,
	\begin{align}
		{\rm Pr}\left(\sup_\theta \sup_{r_{ij, \theta}} \left| P(r_{ij, \theta}) - \hat{P}(r_{ij, \theta}) \right| \ge A_{ij}' \sqrt{\frac{\tau}{N}} + B_{ij}' \frac{\tau}{N} \right) \le \exp\left(-\tau\right) .
	\end{align}
\end{lemma}
\paragraph{Proof}
	The proof directly follows by applying Talagrand's inequality~\citep{steinwart2008support}: for any $\tau > 0$,
	\begin{align*}
		{\rm Pr}\left(\sup_\theta \sup_{r_{ij, \theta}} \left| P(r_{ij, \theta}) - \hat{P}(r_{ij, \theta}) \right| \ge \sqrt{\frac{2 \tau (\sigma_{ij}^2 + F_{ij}^2)}{N}} + \frac{2 \tau F_{ij}}{3 N} \right) \le \exp\left(-\tau\right) ,
	\end{align*}
	where $\sigma_{ij}^2 \ge \mathbb{E}[p(x_i, x_j)^2]$ and $F_{ij} \ge \| p(x_i, x_j) \|_\infty$.
	By setting $A_{ij}' = \sqrt{2 (\sigma_{ij}^2 + F_{ij}^2)}$ and $B_{ij}' = 2 F_{ij} / 3$, we obtain the claim.
\hfill $\Box$

\subsection{Proofs}

\paragraph{Proof of Theorem~\ref{th:solution}: }
	Recall that
	\begin{align*}
		\sum_{i, j \in S^\comp} H_{ij} \ge 0 ,
	\end{align*}
	holds for any $S^\comp \subseteq [D]$ from $H \in \mathbb{R}_+^{D \times D}$.
	It is, therefore, sufficient to prove that
	\begin{align}
		\sum_{i, j \in {S^*}^\comp} H_{ij} = \underbrace{\sum_{i \in {S^*}^\comp} H_{ii}}_{\rm (A)} + \underbrace{\sum_{i, j \in {S^*}^\comp: i \neq j} H_{ij}}_{\rm (B)} = 0 .
	\end{align}
	Because $p_i = q_i$ is required for $i \in {S^*}^\comp$ from Condition~(\ref{eq:problem1}), $H_{ii} = g(p_i, q_i) = 0$ must hold for $i \in {S^*}^\comp$, which results in ${\rm (A)} = 0$.
	Similarly, $H_{ij} = g(p_{ij}, q_{ij}) = 0$ is required for $i, j \in {S^*}^\comp$, and we have ${\rm (B)} = 0$.
\hfill $\Box$

\paragraph{Proof of Theorem~\ref{th:hoeffding}: }
	Recall that ${\rm KS}(p_{ij, \theta}, q_{ij, \theta}) \in [0, 1]$ for any $\theta \in [0, \pi]$.
	By applying Hoeffdings's inequality, the claim follows.
\hfill $\Box$

\paragraph{Proof of Theorem~\ref{th:consistent}: }
	The theorem is true for $k = D$ and $k = 0$ because $\hat{S}^\comp = {S^*}^\comp = [D]$ and $\hat{S}^\comp = {S^*}^\comp = \emptyset$ hold, respectively.
	Therefore, we only need to consider the case when $1 \le k \le D-1$.
	
	Let $\epsilon = \vi H - \hat{H} \vi_\infty$ and $f(T, T'; M) = \sum_{i \in T, j \in T'} M_{ij}$ for a matrix $M$.
	We also define the index sets $U = {S^*}^\comp \setminus \hat{S}^\comp$, $V = \hat{S}^\comp \setminus {S^*}^\comp$, and $W = {S^*}^\comp \cap \hat{S}^\comp$.
	We then have
	\begin{align*}
		f({S^*}^\comp, {S^*}^\comp; \hat{H}) - f(\hat{S}^\comp, \hat{S}^\comp; \hat{H}) &= f(U, U; \hat{H}) + 2 f(U, W; \hat{H}) - f(V, V; \hat{H}) - 2 f (V, W; \hat{H}) \\
		& \le f(U, U; H) + 2 f(U, W; H) - f(V, V; H) - 2 f (V, W; H) \\
		& \hspace{12pt} + \epsilon (|U|^2 + |V|^2 + 2 |U| |W| + 2|V| |W|) \\
		& \le f({S^*}^\comp, {S^*}^\comp; H) - f(\hat{S}^\comp, \hat{S}^\comp; H) + 2 \epsilon k^2 \\
		& \le 2 \epsilon k^2 - \eta , 
	\end{align*}
	where, in the first inequality, we used the fact that
	\begin{align*}
		| f(T, T'; H) - f(T, T'; \hat{H}) | \le \epsilon |T| |T'|, 
	\end{align*}
	and in the second inequality, we used
	\begin{align*}
		|U|^2 + |V|^2 + 2 |U| |W| + 2|V| |W| & \le (|U| + |W|)^2 + (|V| + |W|)^2 \\
		& = |{S^*}^\comp|^2 + |\hat{S}^\comp|^2 \\
		& = 2 k^2 .
	\end{align*}
	Recall the assumption $\eta > 0$.
	If $\epsilon \le \eta / 2 k^2$, $f({S^*}^\comp, {S^*}^\comp; \hat{H}) \le f(\hat{S}^\comp, \hat{S}^\comp; \hat{H})$ holds implying ${S^*}^\comp$ is the minimizer of (\ref{eq:kss}), which proves the claim. 
\hfill $\Box$

\paragraph{Proof of Theorem~\ref{th:approx}: }
	The proof directly follows from the fact that the monotone submodular maximization problem with a cardinality constraint is $(1-1/e)$- approximable~\citep{nemhauser1978analysis}.
	We note that problem (\ref{eq:kss}) is equivalent to finding an $\hat{S}$ that maximizes $f'(\hat{S})$ under a cardinality constraint.
	The basic assumption $f'(\emptyset) = 0$ of ~\cite{nemhauser1978analysis} is trivial from the definition of $f'$.
	It, therefore, remains to prove that $f'$ is monotone submodular.
	For $A, B \in [D]$, we observe that
	\begin{align*}
		f'(A \cup B) + f'(A \cap B) &= \left( \sum_{i, j \in [D]} \hat{H}_{ij} - \sum_{i, j \in (A \cup B)^\comp} \hat{H}_{ij} \right) + \left( \sum_{i, j \in [D]} \hat{H}_{ij} - \sum_{i, j \in (A \cap B)^\comp} \hat{H}_{ij} \right) \\
		&= \left( \sum_{i, j \in [D]} \hat{H}_{ij} - \sum_{i, j \in A^\comp} \hat{H}_{ij} \right) + \left( \sum_{i, j \in [D]} \hat{H}_{ij} - \sum_{i, j \in B^\comp} \hat{H}_{ij} \right) \\
		& \hspace{12pt} - 2 \sum_{i \in A^\comp \setminus B^\comp, j \in B^\comp \setminus A^\comp} \hat{H}_{ij} \\
		& \le f'(A ) + f'(B) ,
	\end{align*}
	which proves that $f'$ is submodular.
	The monotonicity can be proved as, for $A \subseteq B \subseteq [D]$, 
	\begin{align*}
		f'(B) - f'(A) = \sum_{i, j \in A^\comp \setminus B^\comp} \hat{H}_{ij} + 2 \sum_{i \in A^\comp \setminus B^\comp, j \in B^\comp} \hat{H}_{ij} \ge 0 .
	\end{align*}
\hfill $\Box$

\paragraph{Proof of Lemma~\ref{lem:diag}: }
	The proof follows using the Dvoretzky-Kiefer-Wolfowitz inequality~\citep{dvoretzky1956asymptotic,massart1990tight}: for any $\delta > 0$, 
	\begin{align}
		{\rm Pr} \left( \|P_i - \hat{P}_i\|_\infty > \delta \right) \le 2 \exp \left( - 2 \delta^2 N \right) ,
	\end{align}
	where $P_i$ is the distribution function of $p_i$ and $\hat{P}_i$ is its empirical counterpart.
	From Lemma~\ref{lem:lem1}, we have
	\begin{align*}
		{\rm Pr}\left( \left| {\rm KS}(p_i, q_i) - {\rm KS}(\hat{p}_i, \hat{q}_i) \right| > \delta \right) \le {\rm Pr}\left( \|P_i - \hat{P}_i\|_\infty + \|Q_i - \hat{Q}_i\|_\infty > \delta \right) .
	\end{align*}
	Hence, it follows that
	\begin{align*}
		{\rm Pr}\left( \left| {\rm KS}(p_i, q_i) - {\rm KS}(\hat{p}_i, \hat{q}_i) \right| > \delta \right) & \le {\rm Pr}\left( \|P_i - \hat{P}_i\|_\infty  > \frac{\delta}{2} \right) + {\rm Pr} \left( \|Q_i - \hat{Q}_i\|_\infty > \frac{\delta}{2} \right) \\
		& \le 4 \exp \left( - \frac{\delta^2}{2} N \right) .
	\end{align*}
\hfill $\Box$

\paragraph{Proof of Lemma~\ref{lem:offdiag}: }
	Recall that
	\begin{align*}
		& {\rm Pr}\left(\left| g(p_{ij}, q_{ij}) - \hat{g}_L(\hat{p}_{ij}, \hat{q}_{ij}) \right| > \delta \right) \\
		 & \le {\rm Pr}\left( \left| g(p_{ij}, q_{ij}) - g(\hat{p}_{ij}, \hat{q}_{ij}) \right| + \left| g(\hat{p}_{ij}, \hat{q}_{ij}) - \hat{g}_L(\hat{p}_{ij}, \hat{q}_{ij}) \right| > \delta \right) \\
		 & \le {\rm Pr}\left( \left| g(p_{ij}, q_{ij}) - g(\hat{p}_{ij}, \hat{q}_{ij}) \right| > \frac{\delta}{2}\right) + {\rm Pr}\left(\left| g(\hat{p}_{ij}, \hat{q}_{ij}) - \hat{g}_L(\hat{p}_{ij}, \hat{q}_{ij}) \right| > \frac{\delta}{2} \right) ,
	\end{align*}
	holds.
	We note that the next inequality holds from Theorem~\ref{th:hoeffding}:
	\begin{align*}
		{\rm Pr}\left(\left| g(\hat{p}_{ij}, \hat{q}_{ij}) - \hat{g}_L(\hat{p}_{ij}, \hat{q}_{ij}) \right| > \frac{\delta}{2} \right) \le 2 \exp \left( - \frac{\delta^2}{2} L \right) .
	\end{align*}
	It therefore remains to prove the next inequality:
	\begin{align}
		{\rm Pr}\left( \left| g(p_{ij}, q_{ij}) - g(\hat{p}_{ij}, \hat{q}_{ij}) \right| > \frac{\delta}{2}\right) \le 2  \exp \left( - 2 C_{ij, \delta} N \right) .
		\label{eq:first}
	\end{align}
	From Lemma~\ref{lem:lem2}, we have
	\begin{align*}
		{\rm Pr}\left( \left| g(p_{ij}, q_{ij}) - g(\hat{p}_{ij}, \hat{q}_{ij}) \right| > \frac{\delta}{2} \right) \le {\rm Pr}\left( \sup_{\theta \in [0, \pi]} \left( \|P_{ij, \theta} - \hat{P}_{ij, \theta}\|_\infty + \|Q_{ij, \theta} - \hat{Q}_{ij, \theta}\|_\infty \right) > \frac{\delta}{2} \right) ,
	\end{align*}
	where $P_{ij, \theta}$, $\hat{P}_{ij, \theta}$ and $Q_{ij, \theta}$, $\hat{Q}_{ij, \theta}$  are the true and empirical distribution functions of variable  $r_{ij, \theta} = x_i \cos \theta + x_j \sin \theta$ under the distributions $p$ and $q$, respectively.
	Moreover, from Lemma~\ref{lem:lem3}, there exists $A_{ij}', B_{ij}', A_{ij}'', B_{ij}'' > 0$ such that, for any $\tau > 0$,
	\begin{align*}
		& {\rm Pr}\left( \sup_{\theta \in [0, \pi]} \|P_{ij, \theta} - \hat{P}_{ij, \theta}\|_\infty > A_{ij}' \sqrt{\frac{\tau}{N}} + B_{ij}' \frac{\tau}{N} \right) \le \exp \left( - \tau \right) , \\
		& {\rm Pr}\left( \sup_{\theta \in [0, \pi]} \|Q_{ij, \theta} - \hat{Q}_{ij, \theta}\|_\infty > A_{ij}'' \sqrt{\frac{\tau}{N}} + B_{ij}'' \frac{\tau}{N} \right) \le \exp \left( - \tau \right) .
	\end{align*}
	Hence, we have
	\begin{align*}
		{\rm Pr}\left( \left| g(p_{ij}, q_{ij}) - g(\hat{p}_{ij}, \hat{q}_{ij}) \right| > A_{ij} \sqrt{\frac{\tau}{N}} + B_{ij} \frac{\tau}{N} \right) \le 2 \exp \left( -\tau \right) ,
	\end{align*}
	where $A_{ij} = A_{ij}' + A_{ij}''$ and $B_{ij} = B_{ij}' + B_{ij}''$.
	By solving $\tau$ for $\delta / 2 = A_{ij} \sqrt{\tau / N} + B_{ij} \tau / N$, we obtain the inequality (\ref{eq:first}).
\hfill $\Box$

\paragraph{Proof of Theorem~\ref{th:main}: }
	Let $\delta = \eta / 2 k^2$.
	From Theorem~\ref{th:consistent}, we have
	\begin{align*}
		{\rm Pr}(\hat{S} \neq S^*) & \le {\rm Pr} \left( \vi H - \hat{H} \vi_\infty > \delta \right) \\
		& \le \sum_{i \in [D]} {\rm Pr} \left( \left| g(p_i, q_i) - g(\hat{p}_i, \hat{q}_i) \right| > \delta \right) \\
		& \hspace{12pt} + \sum_{i, j \in [D] : i > j} {\rm Pr} \left(\left| g(p_{ij}, q_{ij}) - \hat{g}_L(\hat{p}_{ij}, \hat{q}_{ij}) \right| > \delta \right) \\
		& \le \sum_{i \in [D]} 4 \exp \left( - \frac{\delta^2}{2} N \right) \\
		& \hspace{12pt} + \sum_{i, j \in [D] : i > j} \left\{ 2 \exp \left( - 2 C_{ij, \delta} N \right) + 2 \exp \left( - \frac{\delta^2}{2} L \right) \right\} \\
		& \le 4 D \exp \left( - \frac{\delta^2}{2} N \right) + D (D - 1) \left\{ \exp \left( - 2 C_\delta N \right) + \exp \left( - \frac{\delta^2}{2} L \right) \right\} , 
	\end{align*}
	where we used Lemma~\ref{lem:diag} and Lemma~\ref{lem:offdiag} in the second inequality and $C_\delta = \min_{i, j \in [D] : i > j} C_{ij, \delta}$. \\
\hfill $\Box$

\paragraph{Proof of Corollary~\ref{cor:n}: }
	To guarantee ${\rm Pr(\hat{S} \neq S^*)} \le \epsilon$, we bound each term of (\ref{eq:main}) as
	\begin{align*}
		 & 4 D \exp \left( - \frac{\eta^2}{8 k^4} N \right) \le \frac{\epsilon}{3}, \\
		 & D (D - 1) \exp \left( - 2 C_{\eta / 2k^2} N \right) \le \frac{\epsilon}{3}, \\
		 & D (D - 1) \exp \left( - \frac{\eta^2}{8 k^4} L \right) \le \frac{\epsilon}{3} .
	\end{align*}
	From each inequality, we derive
	\begin{align*}
		& N \ge \frac{8 k^4}{\eta^2} \log \frac{12D}{\epsilon} = O\left( \frac{k^4}{\eta^2} \log \frac{D}{\epsilon} \right) , \\
		& N \ge \frac{1}{2 C_{\eta / 2k^2}} \log \frac{3 D (D - 1)}{\epsilon} = O\left( \frac{1}{C_{\eta/2k^2}} \log \frac{D}{\epsilon} \right), \\
		& L \ge \frac{8 k^4}{\eta^2} \log \frac{3 D (D - 1)}{\epsilon} = O\left( \frac{k^4}{\eta^2} \log \frac{D}{\epsilon} \right) .
	\end{align*}
\hfill $\Box$

\paragraph{Proof of Theorem~\ref{th:consistency_necc}: }
	We prove by contraposition.
	Suppose there exists $a \in S^*$ such that (N1') and (N2') hold:
	\begin{align}
		({\rm N1'}) \; H_{aa} = 0 , \qquad ({\rm N2'}) \; \forall b \in [D] \setminus \{a\} , \; H_{ab} = 0 .
		\label{eq:nn}
	\end{align}
	Then, for any $c \in {S^*}^\comp$, 
	\begin{align*}
		0 & = \sum_{i, j \in {S^*}^\comp} H_{ij} \\
		& = \sum_{i, j \in ({S^*}^\comp \cup \{a \}) \setminus \{c \}} H_{ij} + \underbrace{2 \sum_{i \in {S^*}^\comp \setminus \{c\}} H_{ci} + H_{cc}}_{=0 \; (\because \forall i, j \in {S^*}^\comp, H_{ij} = 0)} - \underbrace{2 \sum_{i \in {S^*}^\comp \setminus \{c\}} H_{ai} - H_{aa}}_{=0 \; (\because {\rm (N1'), (N2')})} \\
		& = \sum_{i, j \in ({S^*}^\comp \cup \{a\}) \setminus \{c\}} H_{ij} ,
	\end{align*}
	holds.
	This shows that $S' = ({S^*}^\comp \cup \{a\}) \setminus \{c\}$ satisfies $|S'^\comp| = k$ and $\sum_{i, j \in S'^{\rm c}} H_{ij} = \sum_{i, j \in {S^*}^{\rm c}} H_{ij}$, which indicates that $\eta = 0$.
\hfill $\Box$

\paragraph{Proof of Theorem~\ref{th:consistency_suff}: }
	For any $S' \neq S^*$, let $U = {S^*}^\comp \setminus S'^\comp$, $V = S'^\comp \setminus {S^*}^\comp$, and $W = {S^*}^\comp \cap S'^\comp$.
	It then holds that
	\begin{align*}
		\sum_{i, j \in S'^\comp} H_{ij} &= \sum_{i, j \in {S^*}^\comp} H_{ij} + \underbrace{2 \sum_{i \in W, j \in V} H_{ij} + \sum_{i, j \in V} H_{ij}}_{> 0 \; (\because {\rm (S1), (S2)}) } - \underbrace{2 \sum_{i \in W, j \in U} H_{ij} - \sum_{i, j \in U} H_{ij}}_{=0 (\because \forall i, j \in {S^*}^\comp, H_{ij} = 0)} \\
		& > \sum_{i, j \in {S^*}^\comp} H_{ij} ,
	\end{align*}
	which indicates that $\eta > 0$.
\hfill $\Box$

\paragraph{Proof of Theorem~\ref{th:consistency_gauss}: }
	We prove by contraposition.
	Let $p(\bm{x}) := \mathcal{N}(\bm{\mu}, \Sigma)$ and $q(\bm{x}) := \mathcal{N}(\bm{\nu}, \Gamma)$.
	Suppose there exists $a \in S^*$ such that both (N1') and (N2') in (\ref{eq:nn}) hold.
	Condition $({\rm N1}')$ is equivalent to $p(x_a) = q(x_a)$, which implies that
	\begin{align*}
		\mu_a = \nu_a, \qquad \Sigma_{aa} = \Gamma_{aa} .
	\end{align*}
	Similarly, Condition $({\rm N2}')$ is equivalent to $p(x_a, x_b) = q(x_a, x_b)$ for any $b \in [D] \setminus \{a\}$, which implies
	\begin{align*}
		\Sigma_{ab} = \Gamma_{ab} .
	\end{align*}
	From these results, we have
	\begin{align*}
		p(\bm{x}_{({S^*} \setminus \{a\})^\comp}) = q(\bm{x}_{({S^*} \setminus \{a\})^\comp}) ,
	\end{align*}
	which contradicts with Condition (\ref{eq:problem2}).
	Hence, there exists no $a \in S$ that satisfies Conditions (N1') and (N2').
\hfill $\Box$

\paragraph{Proof of Proposition~\ref{prop:rel}: }
For the bivariate KL-divergence, under the specified conditions,
\begin{align*}
	{\rm KL}[p_{ij} || q_{ij}] & = \frac{1}{2} \left\{ \frac{2 - 2 \Sigma_{ij} \Gamma_{ij}}{1 - \Gamma_{ij}^2} - \log \frac{1 - \Sigma_{ij}^2}{1 - \Gamma_{ij}^2} - 2\right\} \\
	&= \frac{1}{2} \left\{\frac{(\Sigma_{ij} - \Gamma_{ij})^2}{1 - \Gamma_{ij}^2} + \frac{1 - \Sigma_{ij}^2}{1 - \Gamma_{ij}^2} - \log \frac{1 - \Sigma_{ij}^2}{1 - \Gamma_{ij}^2} - 1 \right\} \\
	&\geq \frac{1}{2}\frac{(\Sigma_{ij} - \Gamma_{ij})^2}{1 - \Gamma_{ij}^2} \geq \frac{1}{2}|\Sigma_{ij} - \Gamma_{ij}| - \frac{1}{8} ,
\end{align*}
holds, where we used the assumption that $\Sigma$ and $\Gamma$ are invertible which implies $\Sigma_{ij}^2, \Gamma_{ij}^2 < 1$, and $t - \log t \geq 1$ for $t > 0$.
\hfill $\Box$

\vskip 0.2in
\bibliography{main}

\end{document}